\documentclass[11pt]{article}

\usepackage[final]{acl}

\usepackage{times}
\usepackage{latexsym}

\usepackage[T1]{fontenc}

\usepackage{hyperref}
\usepackage{url}
\usepackage{multirow}
\usepackage{algorithm}
\usepackage{algpseudocode}
\usepackage{graphicx} 
\usepackage{float}    
\usepackage{tikz}
\usetikzlibrary{shapes,arrows}
\usepackage{subcaption} 
\usepackage{tabularx}
\usepackage{pifont}
\usepackage{booktabs}
\usepackage{verbatim}
\usepackage{tablefootnote}
\usepackage[normalem]{ulem}

\usepackage{xcolor}
\usepackage{tcolorbox}
\tcbuselibrary{breakable, listings, skins}

\newtcolorbox{examplebox}[2][]{
  colback=white,
  colframe=black!60,
  title=#2,
  fonttitle=\bfseries,
  sharp corners,
  boxrule=0.6pt,
  width=\columnwidth,
  fontupper=\small,
  #1
}

\usepackage{listings}

\usepackage{arydshln} 

\usepackage{amsmath,amssymb}

\usepackage{cleveref}
\usepackage[utf8]{inputenc}
\usepackage{microtype}
\usepackage{inconsolata}

\title{Failure Modes in Multi-Hop QA: The Weakest Link Effect and the Recognition Bottleneck}

\author{Meiru Zhang \\
  University of Cambridge \\
  \texttt{mz468@cam.ac.uk} \\
\And
  Zaiqiao Meng \\
  University of Cambridge \\
  \texttt{zm324@cam.ac.uk} \\
\And
  Nigel Collier \\
  University of Cambridge \\
  \texttt{nhc30@cam.ac.uk} }

\begin{document}
\maketitle
\begin{abstract}
Despite scaling to massive context windows, Large Language Models (LLMs) struggle with multi-hop reasoning due to inherent position bias, which causes them to overlook information at certain positions. Whether these failures stem from an inability to \textit{locate} evidence (recognition failure) or \textit{integrate} it (synthesis failure) is unclear. We introduce Multi-Focus Attention Instruction (MFAI), a semantic probe to disentangle these mechanisms by explicitly steering attention towards selected positions. Across 5 LLMs on two multi-hop QA tasks (MuSiQue and NeoQA), we identify the ``Weakest Link Effect'': in our 18-document, 3-bucket setting, multi-hop reasoning performance collapses to the level of the least visible evidence, governed by absolute position rather than the linear distance between facts. While matched MFAI resolves recognition bottlenecks, improving accuracy by up to 11.49\% in low-visibility positions, misleading MFAI yields divergent effects modulated by task topology: entity-centric tasks with vertical reasoning chains are vulnerable, whereas event-centric tasks with horizontal evidence structures are more resilient. Finally, we demonstrate that ``thinking'' models utilizing System-2 reasoning effectively locate and integrate the required information, matching gold-only baselines even in noisy, long-context settings. Supplementary experiments on 2WikiMultiHopQA, extended 3--4 hop counts, and a 32B model confirm these findings generalize across datasets, reasoning depths, and model scales.
\end{abstract}

\section{Introduction}
\label{sec:introduction}

While the theoretical capacity of Large Language Models (LLMs) has expanded exponentially, with context windows scaling from 4k to millions of tokens, the \textit{effective} utilization of this context remains fundamentally constrained~\citep{kamradt2023needle, an2025whyeffectivelengthfallshort,hsieh2024ruler}. Previous studies have characterized this limitation as position bias due to attention failures, such as the Lost-in-the-Middle phenomenon~\citep{liu2024lostinthemiddle}, primacy bias~\citep{liu2024lostinthemiddle}, and recency bias~\citep{press2021shortformer,sun2021recencybias}, where information at specific positions is systematically overlooked. This bias is often attributed to model mechanisms such as position embedding~\citep{wang2025eliminatingpositionbiasmechanisticapproach}, attention sinks~\citep{xiao2024efficientstreaming}, where initial tokens monopolize attention mass, and the failure of induction heads~\citep{olsson2022context} to effectively copy information from mid-context positions.

Recent research extends position bias analysis to Multi-Hop Question Answering (MHQA), where models must synthesize disconnected evidence~\citep{yu2025unleashingmultihopthroughmisorderedcontext,huang2025maskinganalyisoflmwithcontextpermutation,baker2024lostinthemiddleandinbetween}. While prior work suggests performance degrades linearly with the distance between facts~\citep{baker2024lostinthemiddleandinbetween,huang2025maskinganalyisoflmwithcontextpermutation} and the number of document splits~\citep{levy2025moredocumentssamelength}, we challenge this view by analyzing evidence topology. By partitioning the context into positional ``buckets'', we observe a step-function behavior: the between-bucket gap on MuSiQue is roughly $4\times$ larger than the within-bucket variation (mean $8.31\%$ vs. $1.87\%$; up to $14.75\%$ for Ministral-8B-Instruct), indicating that attention operates at bucket-level granularity rather than fine-grained distance. Instead, multi-hop reasoning follows a ``Weakest Link Effect'': performance is governed by the absolute bucket position of the least visible evidence. If a single reasoning hop falls into an under-attended region, the entire chain collapses.

Existing position bias mitigation methods are predominantly resource-intensive, relying on finetuning for data augmentation~\citep{li2024makingbettermultihopreasoner, he2024neverlostinthemiddle} or modifying inference compute via mechanistic modifications~\citep{wang2025eliminatingpositionbiasmechanisticapproach,chen2025searchincontext}. In contrast, \citet{zhang2024attentioninstruction} propose Attention Instruction, a training-free intervention that uses natural language to explicitly direct the model's focus. They demonstrated that absolute indexing (e.g., ``Document 1'') can effectively override position bias, allowing retrieval from the typically lost middle. While this validation was limited to one-hop question answering, the ability to steer attention provides a powerful control variable. It allows us to artificially restore evidence visibility, potentially isolating the \textbf{recognition failures} (overlooked evidence) from \textbf{synthesis failures} (inability to connect facts).

Building on these observations, we introduce \texttt{Multi-Focus Attention Instruction (MFAI)}.\footnote{Code and data are available at \url{https://github.com/cambridgeltl/weakest-link-effect}.} MFAI acts as a semantic probe, explicitly indexing evidence locations to simulate successful recognition within a factorial experiment (details in \Cref{sec:experimental-setup}). By comparing performance with and without these instructions, we aim to isolate recognition failures from synthesis failures. This distinction offers insights for Retrieval-Augmented Generation (RAG) architectures~\citep{lewis2020retrieval,yu2024evaluationofrag}.

By deploying MFAI on the two MHQA datasets, MuSiQue~\citep{trivedi2022musique} and NeoQA~\citep{glockner2025neoqa}, we address three research questions: (1) Does linear distance between supporting facts or their absolute position within the context govern multi-hop performance? (2) Can MFAI effectively mitigate position bias in MHQA and isolate recognition errors from intrinsic synthesis limitations? (3) Does extended test-time compute (System-2 reasoning) confer robustness against misleading attention cues?

\paragraph{Contributions}
In summary, this work offers the following contributions:
\begin{itemize}\setlength\itemsep{1pt}
    \item \textbf{The ``Weakest Link Effect'':} We provide a granular analysis of position bias in MHQA, showing that performance collapses to the level of the least visible evidence bucket, regardless of the distance between facts. This pattern generalizes across an additional dataset (2WikiMultiHopQA;~\citealp{ho2020constructing}), longer reasoning chains (3--4 hops), and a 4$\times$ larger model (Qwen2.5-32B;~\citealp{qwen2025qwen25technicalreport}).
    \item \textbf{Mechanistic Disentanglement:} We demonstrate that steering attention with semantic instructions resolves the recognition bottleneck, restoring accuracy in low-visibility positions. This confirms that performance drops are largely recognition-based rather than reasoning-based. 
    \item \textbf{The Differential Impact of Attention Steering:} We reveal that while matched MFAI neutralizes position bias, unmatched (misleading) MFAI yields divergent effects modulated by task topology. Misleading cues degrade entity-centric tasks with vertical reasoning chains (MuSiQue) but have limited impact on event-centric tasks with horizontal evidence structures (NeoQA).
    \item \textbf{System-2 Reasoning Robustness:} We show that models utilizing extended test-time compute (e.g., Qwen3-8B-Think) override context retrieval artifacts, exhibiting superior robustness to both inherent position bias and adversarial prompts.
\end{itemize}


\begin{figure*}[t]
    \centering
    \begin{tikzpicture}[
        doc/.style={draw, rectangle, minimum width=0.4cm, minimum height=0.6cm, fill=white, inner sep=0pt},
        gold/.style={draw, rectangle, minimum width=0.4cm, minimum height=0.6cm, fill=orange!40, inner sep=0pt},
        bad/.style={draw, rectangle, minimum width=0.4cm, minimum height=0.6cm, fill=red!30, inner sep=0pt},
        bucket_border/.style={draw, dashed, gray, rounded corners, inner sep=0.1cm},
        arrow/.style={->, thick, red, dashed, shorten >=1mm, shorten <=1mm},
        font=\scriptsize
    ]

    \def\colOneX{0}
    \def\colTwoX{5.5}
    \def\colThreeX{11}
    
    \def\rowTopY{0}
    \def\rowMidY{-1.4} 
    \def\rowBotY{-2.8} 
    
    \def\docOffset{-1.25} 
    \def\docSpacing{0.5} 
    \def\docStartX{-1.25}

    \node[draw, fill=blue!10, rounded corners, minimum width=10cm, align=center] at (5.5, 1.5) {{\small \textbf{Prompt Structure:} [Task] \textcolor{blue}{[Attn Instr]} [Documents 1..18] [Question]}};

    \node[anchor=south] at (\colOneX, 0.6) {\small \textbf{(a) Spread Test}};

    \node[bucket_border, minimum width=3.5cm, minimum height=0.8cm, label=left:Beginning] at (\colOneX, \rowTopY) {};
    \foreach \i in {0,...,5} { \node[doc] at (\colOneX + \docStartX + \i*\docSpacing, \rowTopY) {}; }

    \node[bucket_border, minimum width=3.5cm, minimum height=1cm, label=left:Middle] at (\colOneX, \rowMidY) {}; 
    \foreach \i in {0,...,5} { \node[doc] at (\colOneX + \docStartX + \i*\docSpacing, \rowMidY) {}; }
    \node[gold] at (\colOneX + \docStartX + 0*\docSpacing, \rowMidY) {}; 
    \node[gold] at (\colOneX + \docStartX + 5*\docSpacing, \rowMidY) {}; 
    \draw[<->, thick, blue] (\colOneX + \docStartX + 0*\docSpacing, \rowMidY-0.5) -- (\colOneX + \docStartX + 5*\docSpacing, \rowMidY-0.5) node[midway, below=-2pt] {Dist varies};

    \node[bucket_border, minimum width=3.5cm, minimum height=0.8cm, label=left:Tail] at (\colOneX, \rowBotY) {};
    \foreach \i in {0,...,5} { \node[doc] at (\colOneX + \docStartX + \i*\docSpacing, \rowBotY) {}; }

    \node[anchor=south] at (\colTwoX, 0.6) {\small \textbf{(b) Cross Test}};

    \node[bucket_border, minimum width=3.5cm, minimum height=0.8cm, label=left:Beginning] at (\colTwoX, \rowTopY) {};
    \foreach \i in {0,...,5} { \node[doc] at (\colTwoX + \docStartX + \i*\docSpacing, \rowTopY) {}; }
    \node[gold] at (\colTwoX + \docStartX + 2*\docSpacing, \rowTopY) {};
    \node[below=1pt] at (\colTwoX + \docStartX + 2*\docSpacing, \rowTopY-0.35) {idx $k$};

    \node[bucket_border, minimum width=3.5cm, minimum height=0.8cm, label=left:Middle] at (\colTwoX, \rowMidY) {};
    \foreach \i in {0,...,5} { \node[doc] at (\colTwoX + \docStartX + \i*\docSpacing, \rowMidY) {}; }

    \node[bucket_border, minimum width=3.5cm, minimum height=0.8cm, label=left:Tail] at (\colTwoX, \rowBotY) {};
    \foreach \i in {0,...,5} { \node[doc] at (\colTwoX + \docStartX + \i*\docSpacing, \rowBotY) {}; }
    \node[gold] at (\colTwoX + \docStartX + 2*\docSpacing, \rowBotY) {};
    \node[below=1pt] at (\colTwoX + \docStartX + 2*\docSpacing, \rowBotY-0.35) {idx $k$};

    \node[anchor=south] at (\colThreeX, 0.6) {\small \textbf{(c) Unmatched Logic}};

    \def\colThreeTopY{0}
    \def\colThreeBotY{-2.2} 

    \node[bucket_border, minimum width=3.5cm, minimum height=0.8cm, label=left:Gold Bucket] (U1_bkt) at (\colThreeX, \colThreeTopY) {};
    \foreach \i in {0,...,5} { \node[doc] at (\colThreeX + \docStartX + \i*\docSpacing, \colThreeTopY) {}; }
    \node[gold] (u_gold) at (\colThreeX + \docStartX + 2*\docSpacing, \colThreeTopY) {};
    \node[below=1pt] at (\colThreeX + \docStartX + 2*\docSpacing, \colThreeTopY-0.35) {Loc idx $k$};

    \node[bucket_border, minimum width=3.5cm, minimum height=0.8cm, label=left:Non-gold Bucket] (U2_bkt) at (\colThreeX, \colThreeBotY) {};
    \foreach \i in {0,...,5} { \node[doc] at (\colThreeX + \docStartX + \i*\docSpacing, \colThreeBotY) {}; }
    \node[bad] (u_bad) at (\colThreeX + \docStartX + 2*\docSpacing, \colThreeBotY) {};
    \node[below=1pt] at (\colThreeX + \docStartX + 2*\docSpacing, \colThreeBotY-0.35) {Loc idx $k$};

    \draw[arrow] (u_gold.east) -- (u_bad.east) node[midway, right=3mm, align=left] {\textbf{Mirroring:}\\Pts to same\\local index};

    \end{tikzpicture}
       \vspace{-1em}
    \caption{Experimental setups shown in vertical columns. (a) \textbf{Spread Test:} Gold documents (orange) are placed in the Middle bucket, and the distance between them varies. (b) \textbf{Cross Test:} Gold documents are split between the Beginning and Tail buckets, maintaining the same local index $k$. (c) \textbf{Unmatched Logic:} This illustrates how a misleading (unmatched) instruction points to non-gold documents (red) in an unselected bucket by mirroring the local index $k$ of the true gold documents.}
    \label{fig:main_figure}
 \vspace{-1em}
\end{figure*}

\section{Related Work}
\label{sec:related-work}

\paragraph{Multi-Hop Reasoning and Retrieval-Augmented Challenges.}
Benchmarks for multi-hop question answering evolved from entity-centric tasks like HotpotQA~\citep{yang2018hotpotqa} and MuSiQue~\citep{trivedi2022musique} to complex challenges requiring event bridging~\citep{li2024meqa,glockner2025neoqa}. Despite this evolution, LLMs frequently exhibited ``disconnected reasoning'', arriving at answers via shortcuts rather than valid logical chains~\citep{min2019compositional, li2024deceptivesemanticshortcut, schnitzler2024morehopqa}. Recent taxonomies categorized the failures of LLMs on MHQA into two distinct dimensions: evidence selection failures and synthesis failures~\citep{xu2025llmaslogicalreasoner}. \citet{yang2024latentmultihopreasoning} and \citet{song2025demystifyingdeepsearch} validated this split, observing that models often recalled entities successfully yet failed to utilize them for subsequent reasoning. While recent works addressed synthesis via specialized training~\citep{wang2025rare}, long-context robustness remains under-explored.

RAG aims to bridge the knowledge gaps, though retrieved context often requires extensive denoising to handle conflicting evidence~\citep{liu2024howmuchraghelpreasoning, wang2025ragconflictingevidence}. \citet{cuconasu2024powerofnoise} revealed the ``noise paradox'', where distractors paradoxically improved model performance. LLM-based interventions, such as generating sequential reasoning notes~\citep{yu2024chainofnote} or re-structuring raw information~\citep{li2025structrag}, have shown promise in improving robustness of LLMs. We extend this line of work by using MFAI to mechanically isolate how evidence topology interacts with recognition and synthesis failures.

\paragraph{Mechanisms and Mitigation of Position Bias.}
Position bias (e.g., Lost-in-the-Middle, recency bias) represents a systematic failure in RAG, where models fail to utilize retrieved information located in certain positions~\citep{liu2024lostinthemiddle, wang2023primacy, sun2021recencybias}. Mechanistically, this bias is linked to attention sinks~\citep{xiao2024efficientstreaming} and the decay of induction head efficacy in deep context~\citep{olsson2022context}. While prior studies modeled performance degradation as a linear decay relative to the distance between supporting facts~\citep{baker2024lostinthemiddleandinbetween, huang2025maskinganalyisoflmwithcontextpermutation}, our analysis challenges this continuous view, proposing a discrete step-wise ``Weakest Link Effect''. 

Existing mitigation strategies predominantly relied on training data augmentation~\citep{li2024makingbettermultihopreasoner, he2024neverlostinthemiddle}, architectural modification~\citep{chen2023extendingcontextwindow, zhang2024foundinthemiddle, wang2025eliminatingpositionbiasmechanisticapproach, adiga2025attentionspeaksvolumes, yu2025mitigateviascalinghiddenstates}, or inference-time reordering~\citep{yu2025unleashingmultihopthroughmisorderedcontext, yi2025attentionbasin}. In contrast, we adopt a training-free semantic steering approach. By building on Attention Instruction~\citep{zhang2024attentioninstruction}, a natural language prompt to anchor attention towards a single document, we introduce MFAI to explicitly probe LLMs and mitigate position bias.

\section{Experimental Setup}
\label{sec:experimental-setup}
We design a factorial experiment to disentangle position bias mechanisms using a fixed context of $N=18$ documents, denoted as $\mathcal{D} = \{d_0, \dots, d_{17}\}$. The context is partitioned into three virtual buckets of six documents each: \textbf{Beginning} ($d_0$ to $d_5$), \textbf{Middle} ($d_6$ to $d_{11}$), and \textbf{Tail} ($d_{12}$ to $d_{17}$). We enforce a two-gold document set ($|\mathcal{G}| = 2$) to isolate the effects of absolute position and inter-gold-document distance.


\subsection{Multi-Focus Attention Instruction}
\label{sec:multifocus_attninstr}

Building on the single-document attention steering~\citep{zhang2024attentioninstruction}, we implement MFAI to probe multi-hop scenarios. MFAI explicitly directs model focus to two documents, $d_X$ and $d_Y$, using the template: ``The answer is in Document $X$ and Document $Y$. Use the information from Document $X$ and Document $Y$ as the main reference.'' We design three experimental conditions based on the relationship between the instructed indices $\{X, Y\}$ and the gold set $\mathcal{G}$:

\begin{itemize}\setlength\itemsep{0pt}
    \item \textbf{No MFAI (NA):} The baseline condition with no MFAI, requiring the model to perform both recognition and synthesis of evidence unaided.
    \item \textbf{Matched MFAI:} The instruction references the true global gold document indices ($X, Y \in \mathcal{G}$), simulating successful recognition.
    \item \textbf{Unmatched MFAI:} The instruction deliberately points to indices in a non-gold bucket ($X, Y \notin \mathcal{G}$) to test robustness against misleading recognition signals. We select adversarial indices by \textbf{mirroring the local bucket index} of the gold documents in a non-gold bucket (\Cref{fig:main_figure}(c)), isolating cross-bucket position bias from local-index effects.
\end{itemize}

\paragraph{Diagnostic Scope.} MFAI is designed as a diagnostic probe rather than a deployable technique: it relies on oracle knowledge of gold document positions to create controlled experimental conditions. We note that directing the model to gold documents not only steers attention but also implicitly reduces the effective search space from 18 candidates to 2 highlighted documents, potentially easing synthesis as well. The performance gain from Matched MFAI therefore provides an \textit{upper-bound estimate} of the recognition bottleneck rather than a precise isolation of it. We use the gold-only ablation (\Cref{appx:ablation}), where models receive only the two gold documents with no distractors, as a ceiling reference that quantifies the remaining gap attributable to search-space difficulty.

\subsection{Research Questions}
\label{sec:research-questions}

We investigate three core questions:
\begin{itemize}
    \item \textbf{RQ1 (Topology):} Does the linear distance between distributed facts or the absolute position govern multi-hop reasoning? We address this by comparing the \textbf{Spread protocol} (distance variation within buckets) against the \textbf{Cross protocol} (split across buckets), as detailed in \Cref{sec:spread-cross-protocols}.
    \item \textbf{RQ2 (Mechanism):} Are failures driven by recognition deficits or intrinsic synthesis limitations? We probe this by applying \textbf{Matched MFAI} to isolate recognition failures from synthesis failures.
    \item \textbf{RQ3 (Robustness):} Does System-2 reasoning improve robustness against misleading signals? We test \textbf{Unmatched MFAI} (misleading cues) on standard instruction-following models vs. thinking models (e.g., Qwen3-8B-Think) to evaluate test-time verification capabilities.
\end{itemize}

\subsection{Spread and Cross Protocols}
\label{sec:spread-cross-protocols}

To characterize reasoning failures, we employ two topological protocols (visualized in \Cref{fig:main_figure}). We evaluate both protocols under all three MFAI conditions (No MFAI, Matched, Unmatched) and report the \textbf{Unmatched} performance as the average accuracy across specific adversarial variants (detailed in \Cref{appx:unmatched_variants}).

\paragraph{Spread Test (Distance within Bucket):} As shown in \Cref{fig:main_figure}(a), we fix one gold document at the bucket start (local idx 0) and vary the second gold document within the same bucket. This tests the model's ability to synthesize information over varying inter-gold-document distances while holding the absolute bucket position constant.

\paragraph{Cross Test (Split across Buckets):} As shown in \Cref{fig:main_figure}(b), we place gold documents in \textit{different} buckets (e.g., Beginning and Tail) while maintaining the same local index $k$. This tests the impact of crossing bucket boundaries.


\subsection{Datasets} 
\label{sec:dataset}

We evaluate our methods on two MHQA benchmarks representing distinct reasoning paradigms: MuSiQue~\citep{trivedi2022musique} and NeoQA~\citep{glockner2025neoqa} (examples in \Cref{appx:dataset_examples}). MuSiQue is an entity-based benchmark with questions derived from real-world Wikipedia documents, requiring reasoning to bridge connections between named entities. It serves as a proxy for standard RAG tasks. In contrast, NeoQA is a fully synthesized dataset constructed from fictional event timelines. It demands complex reasoning over entities involved in multiple sequential events, effectively eliminating parametric knowledge shortcuts. 
We filter for questions requiring exactly two gold documents and use a fixed context size of 18 documents (2 gold, 16 distractors). The final evaluation set comprises 1,246 MuSiQue examples (evaluated via Exact Match (EM)) and 402 NeoQA examples (evaluated via Accuracy (Acc.)). We further validate on the Compositional and Inference subsets of 2WikiMultiHopQA~\citep{ho2020constructing}, MuSiQue 3-hop and 4-hop subsets, and NeoQA with alternative distractor settings (\Cref{appx:supplementary-validation}).

\begin{figure*}[!t]
    \centering
    \begin{subfigure}[b]{\textwidth}
        \includegraphics[width=\textwidth]{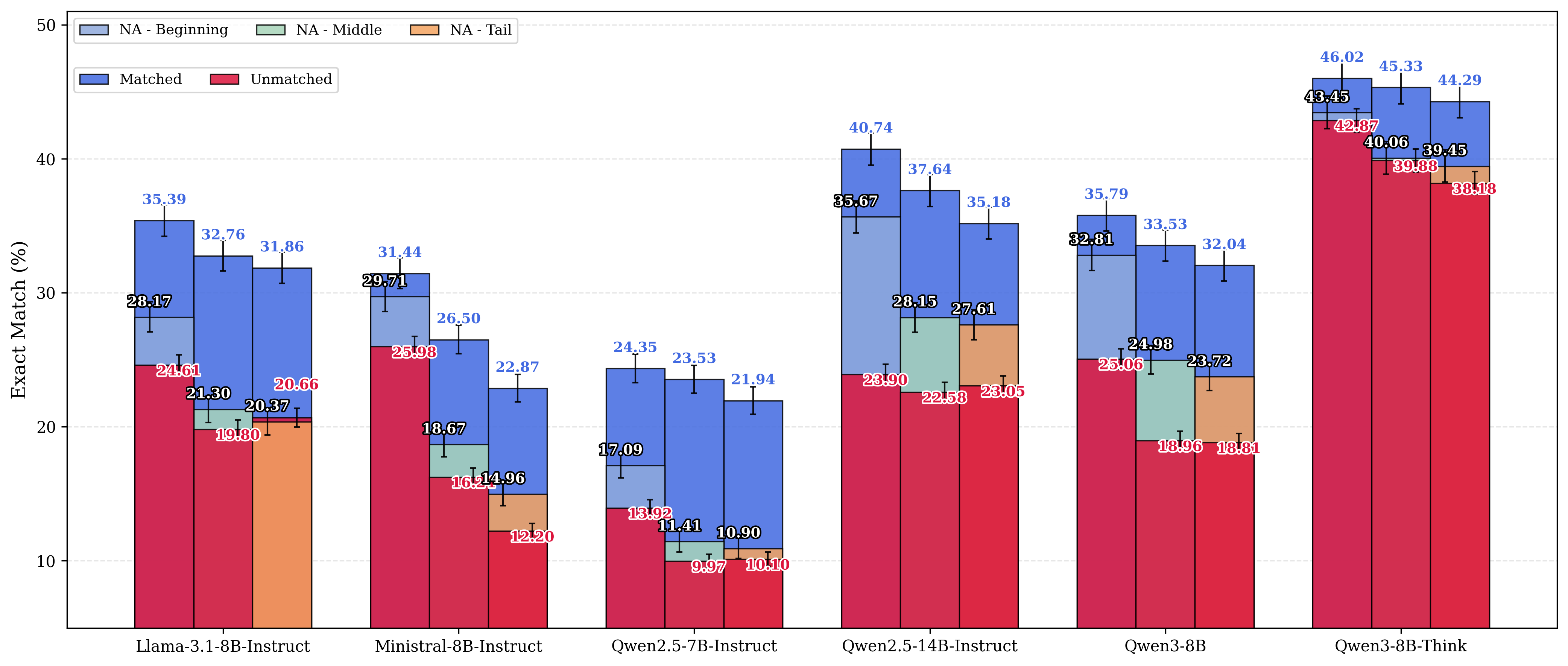}
        \caption{MuSiQue dataset}
    \end{subfigure}
    \begin{subfigure}[b]{\textwidth}
        \includegraphics[width=\textwidth]{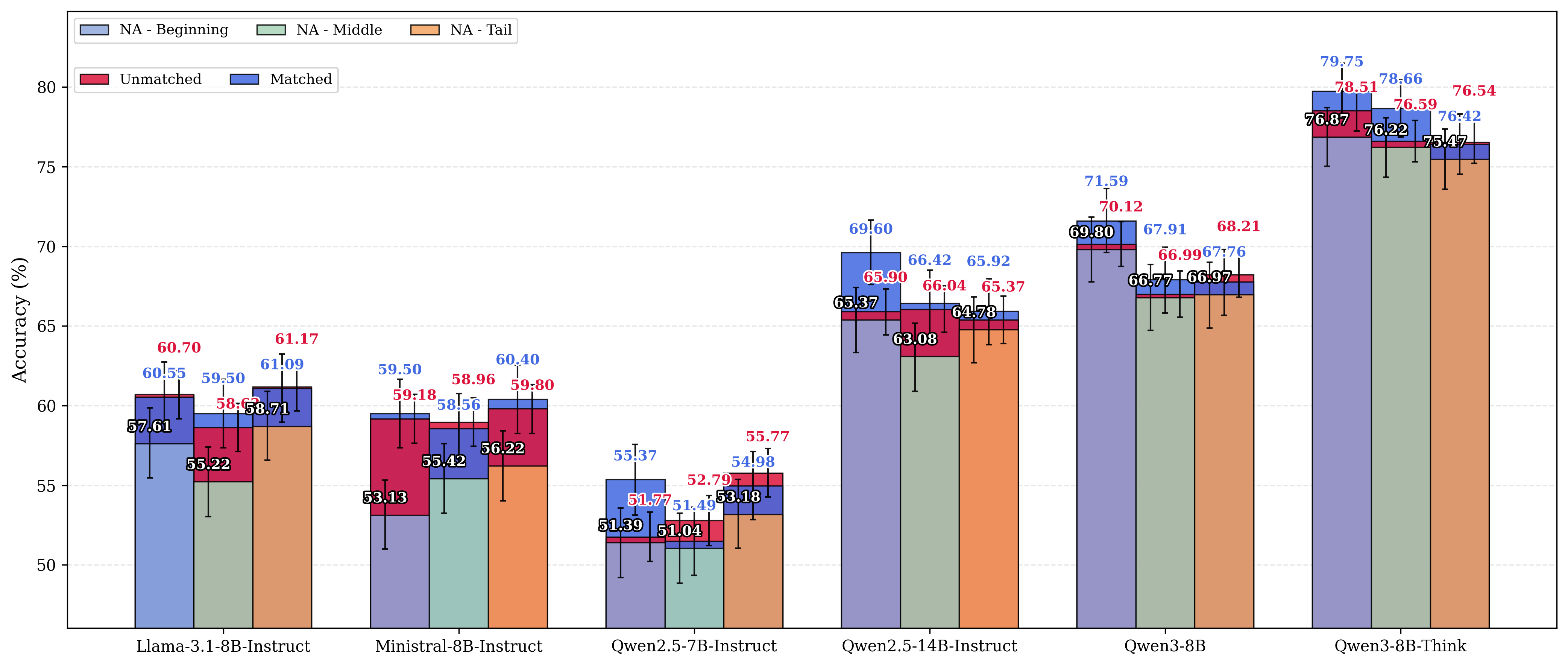}
        \caption{NeoQA dataset}
    \end{subfigure}
    \caption{Model performance across three MFAI conditions (No MFAI, Matched, Unmatched) on MuSiQue and NeoQA in the Spread Test. Bars show mean accuracy (\%) for No MFAI (bucket-colored), Matched (dark blue), and Unmatched (red), pooled across five distances within each bucket. Error bars are 95\% bootstrap CIs. Unmatched averages over adversarial variants (\Cref{appx:unmatched_variants}).}
    \vspace{-3mm}
    \label{fig:spread_test_main}
\end{figure*}

\subsection{Models}
\label{sec:models}

We evaluate five state-of-the-art Large Language Models: Qwen2.5-7B-Instruct~\citep{qwen2025qwen25technicalreport}, Qwen2.5-14B-Instruct~\citep{qwen2025qwen25technicalreport}, Llama-3.1-8B-Instruct~\citep{llama3.1_8b_instruct}, Ministral-8B-Instruct~\citep{ministral_8b_instruct_2410}, and Qwen3-8B~\citep{yang2025qwen3}. Qwen3-8B is assessed in both thinking (triggered via \texttt{<think>}) and non-thinking (triggered via \texttt{</no\_think>}) modes to determine the impact of test-time compute on robustness against position bias and misleading MFAI cues. All models are evaluated at a temperature of 0.0 for reproducibility, consistent with prior work~\citep{levy2025moredocumentssamelength}. We additionally evaluate Qwen2.5-32B-Instruct-GPTQ-Int8 to assess scalability (\Cref{appx:supplementary-validation}).

\begin{figure*}[!t]
    \centering
    \begin{subfigure}[b]{\textwidth}
        \includegraphics[width=\textwidth]{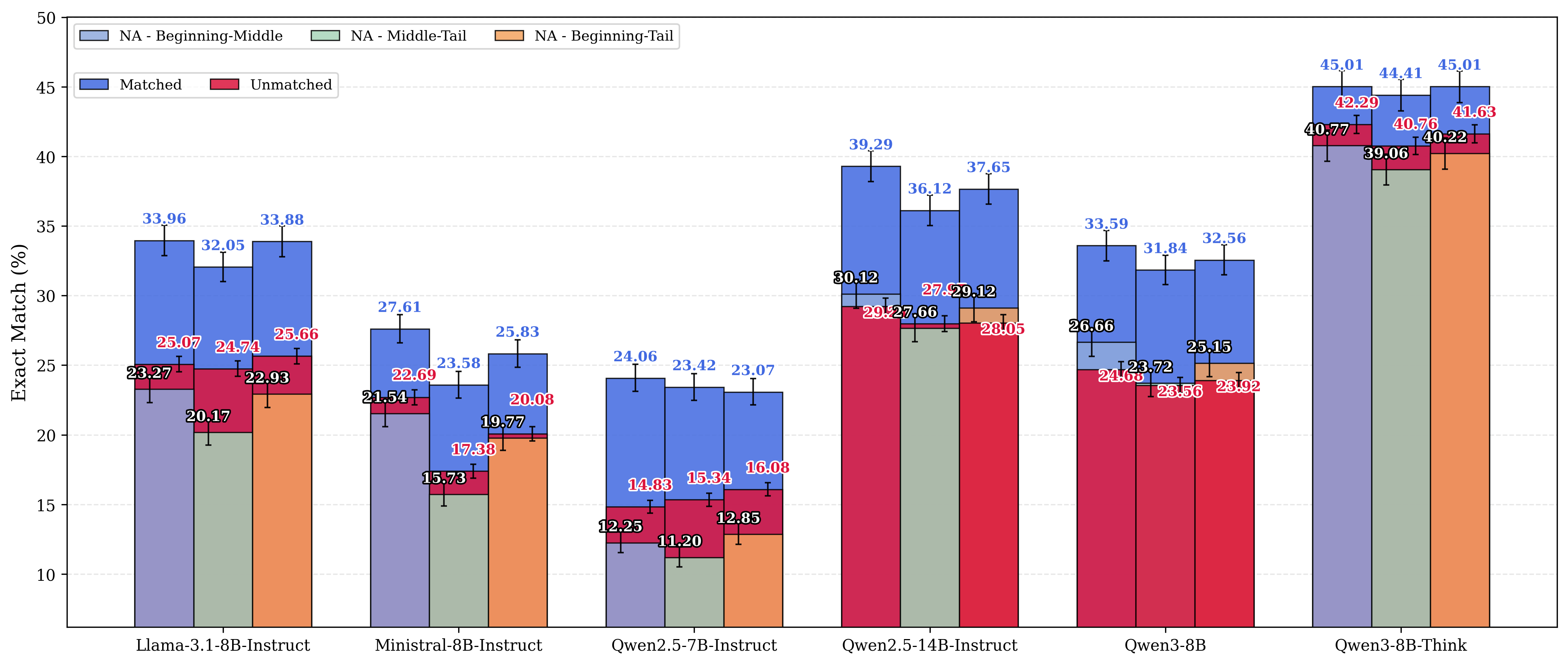}
        \caption{MuSiQue dataset}
    \end{subfigure}
    \begin{subfigure}[b]{\textwidth}
        \includegraphics[width=\textwidth]{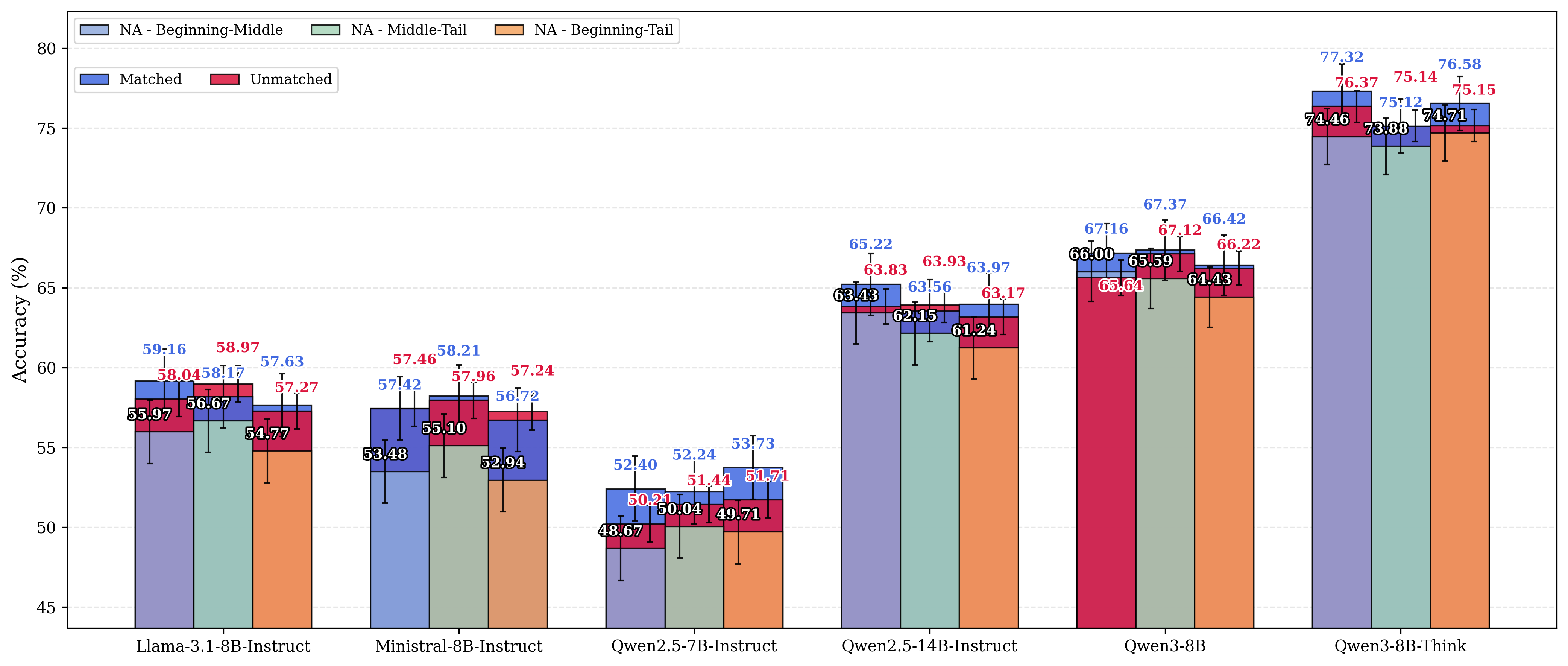}
        \caption{NeoQA dataset}
    \end{subfigure}
    \caption{Model performance across three MFAI conditions (No MFAI, Matched, Unmatched) on MuSiQue and NeoQA in the Cross Test. Bars show mean accuracy (\%) for No MFAI (pair-colored), Matched (dark blue), and Unmatched (red), pooled across six local indices within each bucket pair. Error bars are 95\% bootstrap CIs. Unmatched averages over adversarial variants (\Cref{appx:unmatched_variants}).}
    \label{fig:cross_test_main}
\end{figure*}

\section{Results and Discussion}
\label{sec:results}

\subsection{Evidence Topology: Absolute Position Governs Performance (RQ1)}
\label{sec:bias-distance}

\begin{figure*}[!t]
    \centering
    \begin{subfigure}[b]{\textwidth}
        \includegraphics[width=\textwidth]{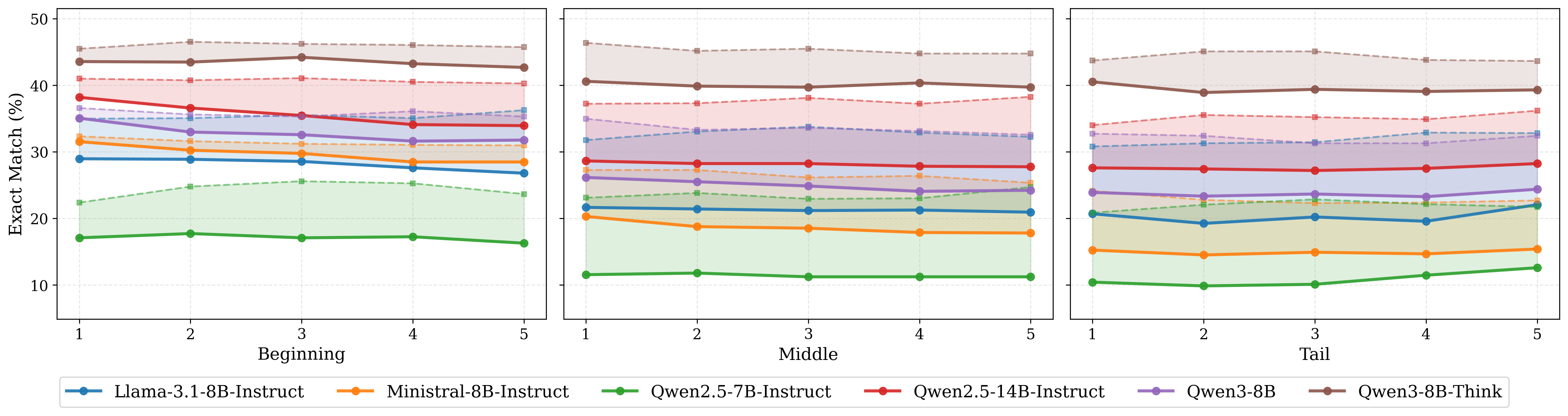}
        \caption{MuSiQue dataset}
    \end{subfigure}
    \begin{subfigure}[b]{\textwidth}
        \includegraphics[width=\textwidth]{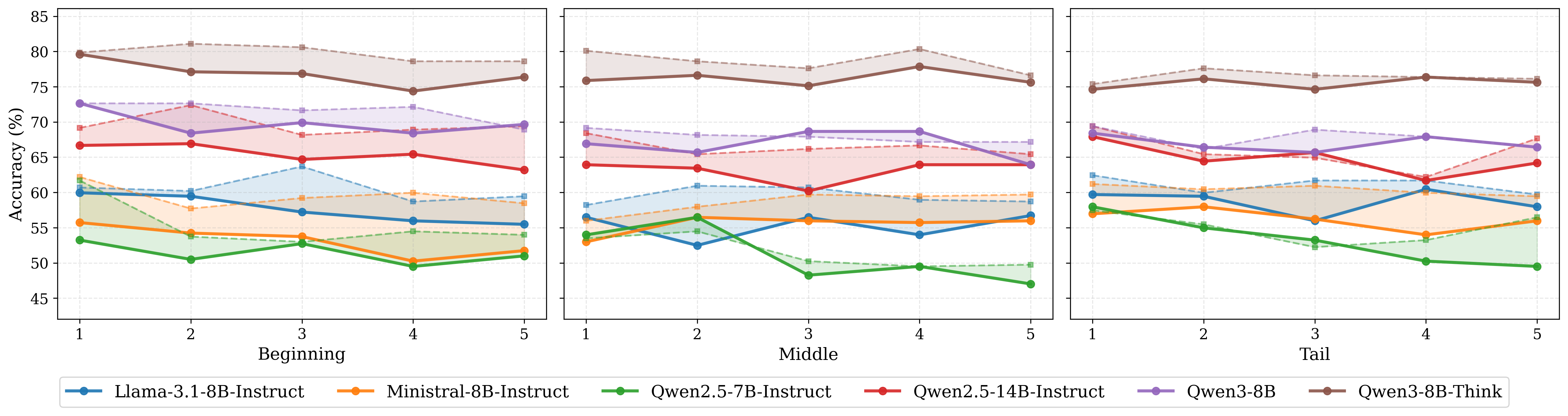}
        \caption{NeoQA dataset}
    \end{subfigure}
    \caption{Model performance across various inter-gold-document distances in each positional bucket. The solid line represents the No MFAI accuracy, whereas the dashed line represents the matched instruction accuracy.}
    \label{fig:spread_test_distance}
\end{figure*}

\begin{figure*}[!t]
    \centering
    \begin{minipage}[b]{0.48\textwidth}
        \centering
        \includegraphics[width=\textwidth]{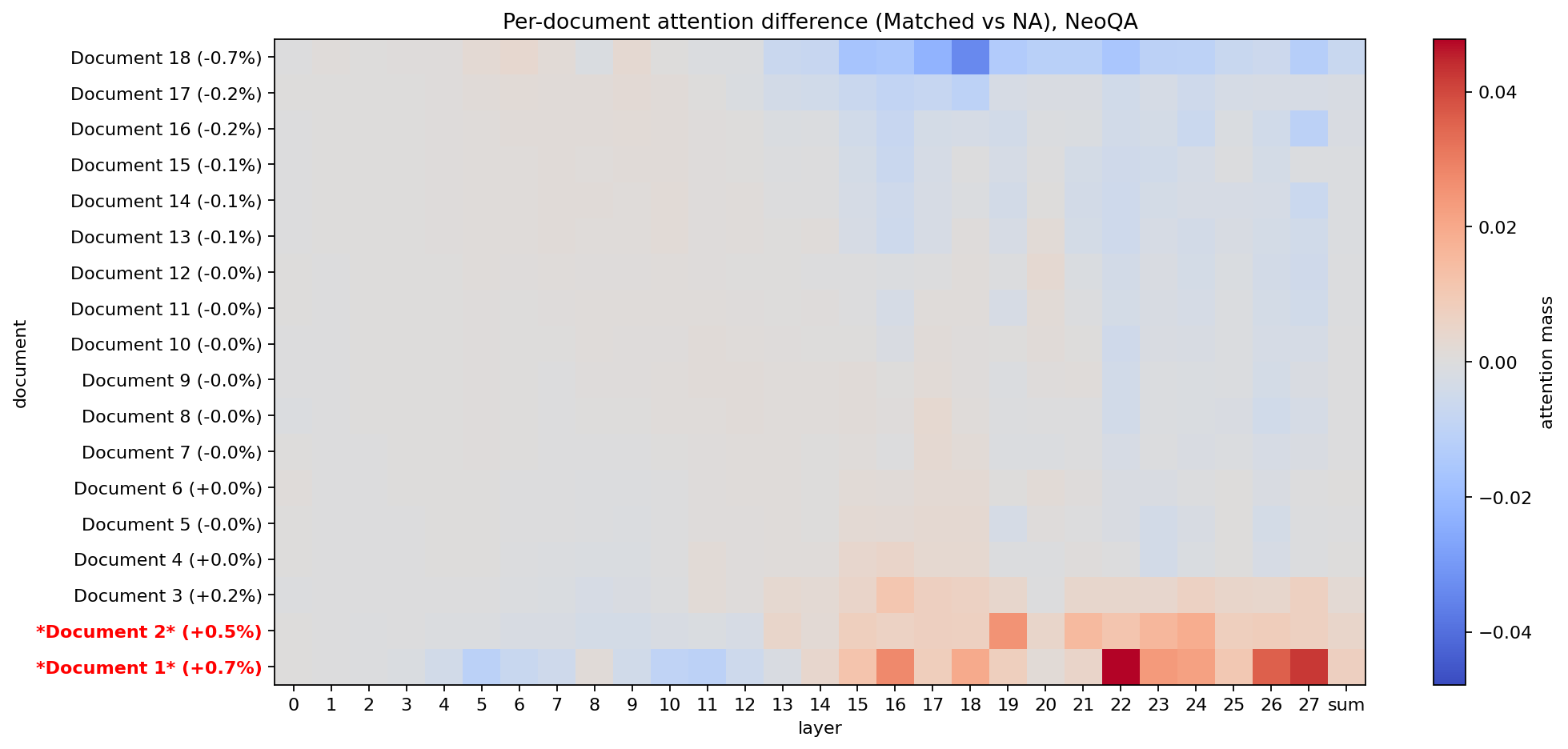}
        \subcaption{Layer-wise Heatmap.}
    \end{minipage}\hfill
    \begin{minipage}[b]{0.48\textwidth}
        \centering
        \includegraphics[width=\textwidth]{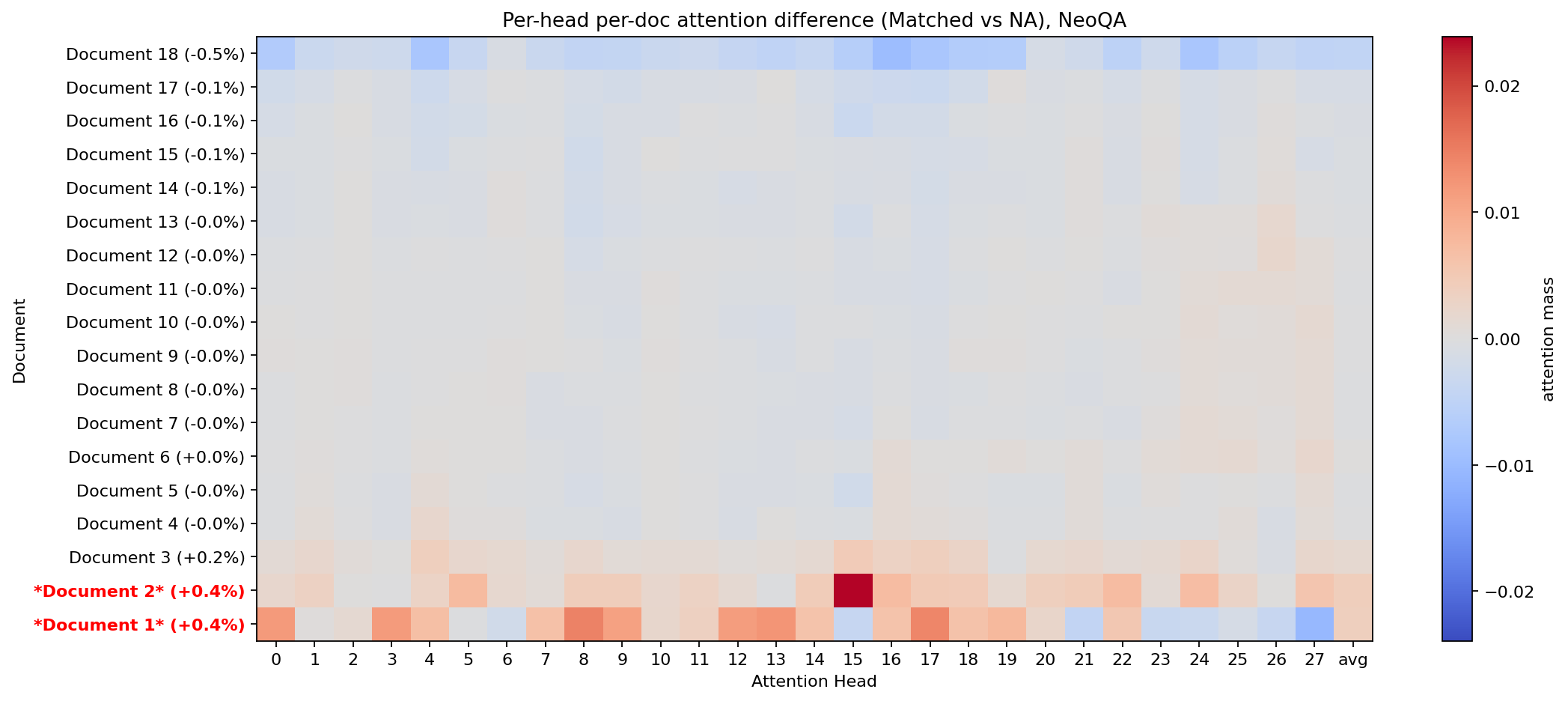}
        \subcaption{Head-wise Heatmap.}
    \end{minipage}
    \caption{Attention mass difference (Matched $-$ No MFAI) of Qwen2.5-7B-Instruct on NeoQA when gold documents are placed in the Beginning bucket (Doc 1 and Doc 2), averaged over 100 examples. The y-axis lists document spans; the x-axis is layer index in (a) and head index in (b). The color bar on the right shows the percentage-point change in normalized attention mass per token. Blue cells indicate a drop; red indicates an increase.}
    \label{fig:attn_heatmap}
\end{figure*}

We first investigate whether multi-hop reasoning performance is determined by the linear distance between supporting facts or their absolute position (RQ1). Focusing on the baseline (No MFAI) results, we observe that position bias is dataset-sensitive: the severity of position bias is much larger on MuSiQue as compared to NeoQA and has different patterns. For example, \Cref{fig:spread_test_main} shows that Ministral-8B-Instruct experiences significant recency bias on MuSiQue but less obvious primacy bias on NeoQA.

\paragraph{Performance follows a Step-Function, not Linear Decay.}
The Spread Test results (\Cref{fig:spread_test_distance}) demonstrate that reasoning performance is largely independent of the distance between evidence documents. Within any fixed bucket (e.g., Beginning), varying the distance between two gold documents from 1 to 5 results in negligible performance variance (typically $\pm 3\%$). For instance, Qwen2.5-14B-Instruct maintains $\sim 28\%$ in the Middle bucket regardless of spread on MuSiQue. However, shifting the entire evidence set from a high-visibility zone (Beginning) to a low-visibility zone (Tail) causes significant drops (e.g., Ministral-8B-Instruct drops by 14.75\% on MuSiQue). This confirms that attention functions as a step-function governed by coarse-grained bucket location. The Cross Test (\Cref{fig:cross_test_main}) shows much smaller variation across the three bucket-pairs (B+M, B+T, M+T) than across single buckets: for example, on MuSiQue, Qwen2.5-14B-Instruct varies by only $2.46\%$ across the bucket-pairs ($30.12\%$, $29.12\%$, $27.66\%$), compared to $8.06\%$ across the corresponding single buckets ($35.67\%$, $28.15\%$, $27.61\%$). Furthermore, \Cref{fig:cross_test_local_idx} shows that performance remains flat across local indices (0 to 5) as long as the bucket-pair is constant, indicating that the model's attentional focus is governed by bucket membership rather than fine-grained offset.

\paragraph{The ``Weakest Link Effect'' of Multi-Hop Reasoning.}
The Cross Test (\Cref{fig:cross_test_main}) further reveals the fragility of multi-hop reasoning. When evidence is split between a high-visibility bucket and a low-visibility bucket, performance does not average out; it collapses toward the weaker bucket. For instance, Ministral-8B-Instruct on MuSiQue achieves $29.71\%$ when both gold documents are in the Beginning bucket and $18.67\%$ when both are in the Middle; with gold split across Beginning+Middle (B+M), accuracy drops to $21.54\%$ — below the $24.19\%$ naive average and close to the weaker Middle rate. We term this the \textbf{Weakest Link Effect}: the probability of successful reasoning is bounded by the minimum recognition probability of any supporting fact. If the model cannot robustly attend to the second hop, the entire reasoning chain breaks, regardless of how salient the first hop was. These step-function and Weakest Link patterns replicate on 2WikiMultiHopQA (\Cref{tab:2wiki-spread-distance}, \Cref{tab:2wiki-cross-local-idx}, \Cref{tab:2wiki-cross-weakest-link}), extend to MuSiQue 3-/4-hop (\Cref{tab:multihop-single-bucket}, \Cref{tab:multihop-cross-weakest-link}), and persist at 32B scale (\Cref{tab:32b-spread-distance}); full results in \Cref{appx:supplementary-validation}.

\subsection{The Nature of Failure: Recognition as the Bottleneck (RQ2)}
\label{sec:nature-of-failure}

Having established where failures occur (position bias), we examine why: are these drops driven by recognition deficits or synthesis limitations?

\begin{figure*}[!t]
    \centering
    \begin{subfigure}[b]{\textwidth}
        \includegraphics[width=\textwidth]{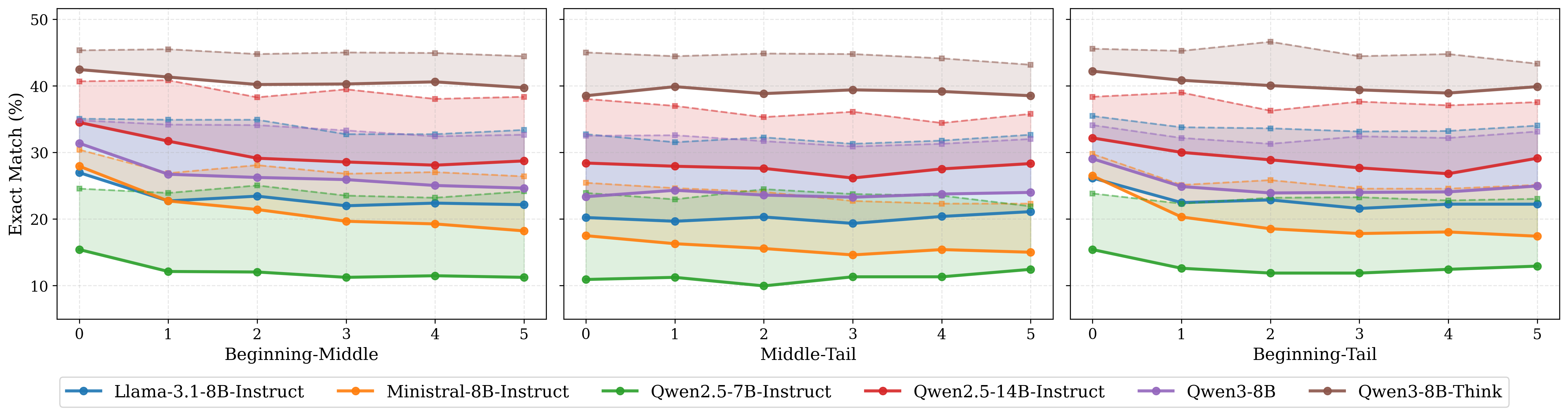}
        \caption{MuSiQue dataset}
    \end{subfigure}
    \begin{subfigure}[b]{\textwidth}
        \includegraphics[width=\textwidth]{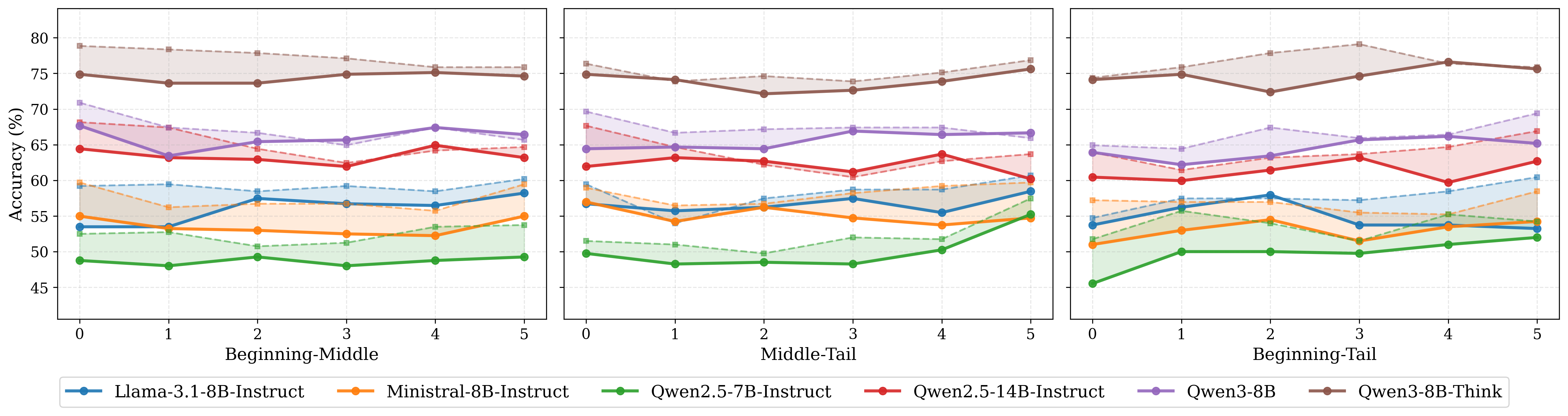}
        \caption{NeoQA dataset}
    \end{subfigure}
    \caption{Model performance across various local indices within the selected pair of positional buckets in the cross test. The solid line represents the No MFAI accuracy, dashed line represents the matched instruction accuracy.}
    \label{fig:cross_test_local_idx}
\end{figure*}

\paragraph{Matched MFAI Restores Performance by Resolving Recognition Deficits.}
We probe this distinction by applying Matched MFAI, which explicitly indexes the gold documents $\mathcal{G}$. For MuSiQue, as shown in \Cref{fig:spread_test_distance}(a) and \Cref{fig:cross_test_local_idx}(a), matched MFAI (dashed lines) consistently boosts the model performance across all inter-gold-document distances and local indices. It elevates performance in the Tail bucket by $4.83\%$ to $11.49\%$ across models on MuSiQue, effectively bridging the gap to Beginning-bucket performance (\Cref{fig:spread_test_main}(a)). In addition, the performance gap between buckets is significantly lower than the baseline (No MFAI, white text) in \Cref{fig:spread_test_main}(a). This provides evidence for the \textbf{recognition bottleneck} hypothesis: models possess the inherent ability to synthesize, but this capacity is bottlenecked by attentional failures.
NeoQA behaves differently: because the baseline position bias is limited, Matched MFAI provides smaller gains, primarily boosting the Beginning bucket on Spread Test (\Cref{fig:spread_test_main}(b)) but shows nearly equal boosting on Cross Test (\Cref{fig:cross_test_main}(b)). This suggests that when models can already effectively reason without bias, Matched MFAI acts as a confirmation signal and further boosts the performance. The Matched MFAI rescue also holds on 2WikiMultiHopQA (\Cref{tab:2wiki-spread-mfai}) and at larger scale on Qwen2.5-32B-Instruct-GPTQ-Int8 (\Cref{tab:32b-spread-mfai}, \Cref{tab:32b-cross-weakest-link}).

\paragraph{Redistributing Attention Mass.}
This recovery of model performance is driven by an observable shift in internal attention. As shown in the attention heatmaps (\Cref{fig:attn_heatmap}), averaged over 100 NeoQA examples where Qwen2.5-7B-Instruct produced a correct answer under Matched MFAI (priority-sampled toward cases in which MFAI causally rescued a No-MFAI failure; see \Cref{appx:heatmap_details}), we observe that attention mass shifts toward the gold documents among all documents (red cells). This increase is uniform across heads but concentrated in deep layers. The same reallocation pattern replicates on MuSiQue (\Cref{fig:attn_heatmap_musique}), confirming that the effect generalizes across multi-hop benchmarks. This redistribution confirms that the failure mode is a recognition bottleneck driven by a lack of attention. The fact that Matched MFAI restores performance implies that the ``compositionality gap''~\citep{press2023measuringandnarrowingcompositionalitygap} is frequently an attention allocation failure rather than a reasoning deficit. McNemar's paired tests~\citep{mcnemar1947note} confirm that this improvement is statistically significant: across all 6 model configurations, 2 datasets, 2 protocols, and 3 positional buckets, 62 of 72 comparisons reach $p < 0.05$ and 56 reach $p < 0.01$ (\Cref{tab:mcnemar-spread,tab:mcnemar-cross}).

\subsection{Robustness and Verification: Unmatched MFAI and System-2 Reasoning (RQ3)}
\label{sec:robustness-verification}

Finally, we investigate model robustness against misleading signals and the role of test-time compute (thinking mode) (RQ3).

\paragraph{Divergent Effects of Unmatched Instructions: Task Topology as a Moderating Factor.}
We observe a distinct difference in how models handle Unmatched MFAI. On MuSiQue, unmatched instructions degrade performance and, for the most instruction-sensitive model Llama-3.1-8B-Instruct, trigger a $\sim 3\times$ increase in response length on the Spread Test (${\sim}331$ to ${\sim}1{,}052$ characters), signaling confusion. Conversely, on NeoQA, unmatched instructions maintain or even slightly improve accuracy. We trace this divergence to the underlying \emph{task topology}. MuSiQue forms a strictly \emph{vertical} causal chain: extracting Entity~A from Document~1 is the prerequisite to searching for Entity~B in Document~2; if attention shifts away from Document~1, the chain breaks entirely. NeoQA, by contrast, is a \emph{horizontal} parallel task in which independent evidence segments (e.g., two dates) can be retrieved separately; even if locally misled, the model can still scan the context to recover both facts. In addition, NeoQA's original distractors come from the same fictional timeline, creating high semantic redundancy that masks positional neglect. When we replaced these same-timeline distractors with documents from random timelines, a clear primacy bias emerged: for Qwen2.5-7B-Instruct, accuracy dropped by 3.83\% from Beginning (62.29\%) to Tail (58.46\%). Matched MFAI reduced this gap to 0.65\%, demonstrating that the attention steering mechanism neutralizes positional bias even in horizontal tasks (\Cref{tab:neoqa-random-timeline}). Detailed analysis of specific unmatched variants (\Cref{appx:detailed-unmatched}) further shows that partially correct instructions (one gold index provided) outperform random non-gold indices, confirming that even imperfect cues can partially resolve the recognition bottleneck.

\paragraph{System-2 Reasoning Overrides the Recognition Artifact.}
Comparing standard instruction-following models with System-2 reasoning models (e.g., Qwen3-8B-Think) reveals that extended test-time compute ($\sim 6\times$ more output tokens than its non-thinking counterpart on NeoQA; see \Cref{tab:think-token-cost}) functions as a verification filter. While standard models passively follow unmatched instructions, the thinking model maintains high accuracy with low variance.

\subsection{Ablation Analysis: The Cost of Noise}
Comparing full-context performance to a gold-only baseline (\Cref{tab:gold_only_ablation} in \Cref{appx:ablation}), standard models (e.g., Llama-3.1-8B) suffer a significant penalty in the 18-document setting (e.g., dropping from $40.45\%$ to avg. $23.28\%$ on MuSiQue), confirming that noise filtering is a primary bottleneck for instruction-following models. In contrast, Qwen3-8B-Think defies this trend, matching or exceeding its gold-only performance even with noise. This suggests that distractors may paradoxically aid thinking models by triggering more rigorous verification. The recognition bottleneck and its rescue by Matched MFAI persist at larger scale on Qwen2.5-32B-Instruct-GPTQ-Int8 (\Cref{tab:32b-spread-mfai}, \Cref{tab:32b-cross-weakest-link}). A no-document ablation (\Cref{tab:no_document_ablation} in \Cref{appx:ablation}) confirms that performance is driven by the provided context rather than parametric memory, as accuracy drops to near-zero when documents are removed.

\section{Conclusion}
\label{sec:conclusion}

We analyzed position bias in long-context multi-hop reasoning, distinguishing between recognition and synthesis deficits. We identified the ``Weakest Link Effect'': multi-hop performance is constrained by the absolute position of the least visible evidence and driven primarily by a recognition bottleneck. This effect replicates across datasets (MuSiQue, NeoQA, 2WikiMultiHopQA), hop counts (2-, 3-, 4-hop), and model scales (7B--32B). Task topology modulates the impact: vertical reasoning chains (MuSiQue) are more vulnerable to misleading attention cues than horizontal tasks (NeoQA), and context homogeneity can mask the underlying bias. Importantly, misleading instructions exhibit a graded effect: partially correct cues (one gold index) still improve over unguided baselines, while fully random cues harm performance, reinforcing that even imperfect recognition signals can partially break the weakest-link bottleneck.

These findings point to several actionable directions. (1)~Since absolute position governs performance, RAG reranking should prioritize high-visibility placement of critical evidence; more broadly, training objectives should target recognition under noisy retrieval, not solely reasoning capabilities. (2)~Developing lightweight mechanistic probes (e.g., attention patterns, logprob signals) would make position bias diagnosis practical without the extensive factorial evaluation our prompt-level approach requires. (3)~Thinking models override position bias at substantial cost (${\sim}6\times$ output tokens; \Cref{tab:think-token-cost}); distilling this self-verification into standard inference remains open. (4)~Whether other reasoning structures (e.g., comparative, temporal) exhibit distinct vulnerability profiles under the vertical--horizontal topology is an open question for task-specific mitigation.

\section*{Limitations}

Our core experiments cover two datasets (MuSiQue, NeoQA) and five models; supplementary evaluations extend to 2WikiMultiHopQA, 3-/4-hop subsets, and Qwen2.5-32B (\Cref{appx:supplementary-validation}), but these use representative models rather than the full suite, and frontier-scale models (70B+) remain untested. We did not perform a mechanistic analysis using logprobs or perplexity, nor explore prompt variations, different bucket counts, or context sizes beyond 18 documents. A pilot study on MFAI index ordering (\Cref{appx:order-sensitivity}) shows negligible sensitivity (median 2.52\% relative change), but was tested on only two models. MuSiQue uses open-ended generation while NeoQA is multiple-choice; the answer options in NeoQA may act as lexical anchors that partially mask position bias independently of task topology, so our attribution of the MuSiQue--NeoQA divergence to topology may be confounded by this format difference. We also report strict Exact Match (EM) for open-ended generation; softer alternatives such as token-level F1 or Contains (normalized substring match) would raise absolute accuracy on MuSiQue by roughly 8--15 percentage points, since they credit sentence-form responses that overlap with (F1) or contain (Contains) the gold answer rather than requiring exact-string equality, though we expect the relative patterns we report to be largely preserved. Finally, we used a fixed distractor order to isolate position bias from semantic similarity effects; real-world scenarios with retrieval rerankers remain future work.

\section*{Ethics Statement}

We affirm that this research follows the ACL Code of Ethics. Our study uses publicly available models and datasets (MuSiQue~\citep{trivedi2022musique} and NeoQA~\citep{glockner2025neoqa}) and does not involve human subjects or private information. By establishing the ``Weakest Link Effect'' to explain position bias in LLMs in multi-hop question answering, our research contributes to the development of more transparent and reliable reasoning systems. We acknowledge the societal risks associated with reasoning LLMs. Our experiments operate within
the existing paradigm of multi-hop reasoning and retrieval-augmented generation. We do not introduce novel risks through our experiments. Code and data are publicly available.\footnote{\url{https://github.com/cambridgeltl/weakest-link-effect}}

We are committed to the full reproducibility of this study. The complete source code and all curated data artifacts are released under a permissive open-source license. Key implementation details and hyperparameters are
described in \Cref{appx:implementation}.

\paragraph{Disclosure of AI Use:} We acknowledge the use of AI assistants for linguistic polishing and grammatical and stylistic revision during the preparation of this manuscript. While these tools were employed to enhance clarity and readability, the authors conducted all experiments and verified all AI-assisted coding. All experimental design, data analysis, scientific reasoning, and final conclusions were produced by the authors, who maintain full responsibility for the content and scientific accuracy of the final paper.

\section*{Acknowledgments}

This work was supported by the Gates Cambridge Scholarship.

\bibliography{custom}

@book{efron1994introduction,
  title={An Introduction to the Bootstrap},
  author={Efron, Bradley and Tibshirani, RJ},
  year={1994},
  publisher={CRC Press}
}

@article{mcnemar1947note,
  title={Note on the sampling error of the difference between correlated proportions or percentages},
  author={McNemar, Quinn},
  journal={Psychometrika},
  volume={12},
  number={2},
  pages={153--157},
  year={1947},
  publisher={Springer-Verlag}
}

@inproceedings{ho2020constructing,
  title={Constructing a multi-hop qa dataset for comprehensive evaluation of reasoning steps},
  author={Ho, Xanh and Nguyen, Anh-Khoa Duong and Sugawara, Saku and Aizawa, Akiko},
  booktitle={Proceedings of the 28th International Conference on Computational Linguistics},
  pages={6609--6625},
  year={2020}
}

@inproceedings{glockner2025neoqa,
    title = "{N}eo{QA}: Evidence-based Question Answering with Generated News Events",
    author = "Glockner, Max  and
      Jiang, Xiang  and
      Ribeiro, Leonardo F. R.  and
      Gurevych, Iryna  and
      Dreyer, Markus",
    editor = "Che, Wanxiang  and
      Nabende, Joyce  and
      Shutova, Ekaterina  and
      Pilehvar, Mohammad Taher",
    booktitle = "Findings of the Association for Computational Linguistics: ACL 2025",
    month = jul,
    year = "2025",
    address = "Vienna, Austria",
    publisher = "Association for Computational Linguistics",
    url = "https://aclanthology.org/2025.findings-acl.616/",
    doi = "10.18653/v1/2025.findings-acl.616",
    pages = "11842--11926",
    ISBN = "979-8-89176-256-5",
}

@inproceedings{li2024meqa,
author = {Li, Ruosen and Wang, Zimu and Tran, Son Quoc and Xia, Lei and Du, Xinya},
title = {MEQA: a benchmark for multi-hop event-centric question answering with explanations},
year = {2024},
isbn = {9798331314385},
publisher = {Curran Associates Inc.},
address = {Red Hook, NY, USA},
booktitle = {Proceedings of the 38th International Conference on Neural Information Processing Systems},
articleno = {4028},
numpages = {28},
location = {Vancouver, BC, Canada},
series = {NIPS '24}
}

@article{trivedi2022musique,
  title={{MuSiQue}: Multihop Questions via Single-hop Question Composition},
  author={Trivedi, Harsh and Balasubramanian, Niranjan and Khot, Tushar and Sabharwal, Ashish},
  journal={Transactions of the Association for Computational Linguistics},
  volume={10},
  pages={539--554},
  year={2022}
}

@inproceedings{levy2025moredocumentssamelength,
    title = "More Documents, Same Length: Isolating the Challenge of Multiple Documents in {RAG}",
    author = "Levy, Shahar  and
      Mazor, Nir  and
      Shalmon, Lihi  and
      Hassid, Michael  and
      Stanovsky, Gabriel",
    editor = "Christodoulopoulos, Christos  and
      Chakraborty, Tanmoy  and
      Rose, Carolyn  and
      Peng, Violet",
    booktitle = "Findings of the Association for Computational Linguistics: EMNLP 2025",
    month = nov,
    year = "2025",
    address = "Suzhou, China",
    publisher = "Association for Computational Linguistics",
    url = "https://aclanthology.org/2025.findings-emnlp.1064/",
    doi = "10.18653/v1/2025.findings-emnlp.1064",
    pages = "19539--19547",
    ISBN = "979-8-89176-335-7"
}

@article{liu2024lostinthemiddle,
  title={Lost in the middle: How language models use long contexts},
  author={Liu, Nelson F and Lin, Kevin and Hewitt, John and Paranjape, Ashwin and Bevilacqua, Michele and Petroni, Fabio and Liang, Percy},
  journal={Transactions of the Association for Computational Linguistics},
  volume={12},
  pages={157--173},
  year={2024}
}

@inproceedings{yu2025unleashingmultihopthroughmisorderedcontext,
  title={Unleashing multi-hop reasoning potential in large language models through repetition of misordered context},
  author={Yu, Sangwon and Kim, Ik-hwan and Song, Jongyoon and Lee, Saehyung and Park, Junsung and Yoon, Sungroh},
  booktitle={Findings of the Association for Computational Linguistics: NAACL 2025},
  pages={6435--6455},
  year={2025}
}

@inproceedings{wang2025eliminatingpositionbiasmechanisticapproach,
  title={Eliminating Position Bias of Language Models: A Mechanistic Approach},
  author={Wang, Ziqi and Zhang, Hanlin and Li, Xiner and Huang, Kuan-Hao and Han, Chi and Ji, Shuiwang and Kakade, Sham M and Peng, Hao and Ji, Heng},
  booktitle={The Thirteenth International Conference on Learning Representations},
  year={2025}
}

@article{zhang2024foundinthemiddle,
  title={Found in the middle: How language models use long contexts better via plug-and-play positional encoding},
  author={Zhang, Zhenyu and Chen, Runjin and Liu, Shiwei and Yao, Zhewei and Ruwase, Olatunji and Chen, Beidi and Wu, Xiaoxia and Wang, Zhangyang and others},
  journal={Advances in Neural Information Processing Systems},
  volume={37},
  pages={60755--60775},
  year={2024}
}

@inproceedings{li2024makingbettermultihopreasoner,
  title={Making long-context language models better multi-hop reasoners},
  author={Li, Yanyang and Liang, Shuo and Lyu, Michael and Wang, Liwei},
  booktitle={Proceedings of the 62nd Annual Meeting of the Association for Computational Linguistics (Volume 1: Long Papers)},
  pages={2462--2475},
  year={2024}
}

@article{baker2024lostinthemiddleandinbetween,
  title={Lost in the Middle, and In-Between: Enhancing Language Models' Ability to Reason Over Long Contexts in Multi-Hop QA},
  author={Baker, George Arthur and Raut, Ankush and Shaier, Sagi and Hunter, Lawrence E and von der Wense, Katharina},
  journal={arXiv preprint arXiv:2412.10079},
  year={2024}
}

@inproceedings{huang2025maskinganalyisoflmwithcontextpermutation,
  title={Masking in multi-hop qa: An analysis of how language models perform with context permutation},
  author={Huang, Wenyu and Vougiouklis, Pavlos and Lapata, Mirella and Pan, Jeff Z},
  booktitle={Proceedings of the 63rd Annual Meeting of the Association for Computational Linguistics (Volume 1: Long Papers)},
  pages={17781--17795},
  year={2025}
}

@inproceedings{sun2021recencybias,
  title={Do Long-Range Language Models Actually Use Long-Range Context?},
  author={Sun, Simeng and Krishna, Kalpesh and Mattarella-Micke, Andrew and Iyyer, Mohit},
  booktitle={Proceedings of the 2021 Conference on Empirical Methods in Natural Language Processing},
  pages={807--822},
  year={2021}
}

@inproceedings{chen2025searchincontext,
  title={Search-in-context: Efficient multi-hop qa over long contexts via monte carlo tree search with dynamic kv retrieval},
  author={Chen, Jiabei and Liu, Guang and He, Shizhu and Luo, Kun and Xu, Yao and Zhao, Jun and Liu, Kang},
  booktitle={Findings of the Association for Computational Linguistics: ACL 2025},
  pages={26443--26455},
  year={2025}
}

@inproceedings{he2024neverlostinthemiddle,
  title={Never lost in the middle: Mastering long-context question answering with position-agnostic decompositional training},
  author={He, Junqing and Pan, Kunhao and Dong, Xiaoqun and Song, Zhuoyang and Liu, Yibo and Sun, Qianguo and Liang, Yuxin and Wang, Hao and Zhang, Enming and Zhang, Jiaxing},
  booktitle={Proceedings of the 62nd Annual Meeting of the Association for Computational Linguistics (Volume 1: Long Papers)},
  pages={13628--13642},
  year={2024}
}

@inproceedings{press2021shortformer,
  title={Shortformer: Better language modeling using shorter inputs},
  author={Press, Ofir and Smith, Noah A and Lewis, Mike},
  booktitle={Proceedings of the 59th Annual Meeting of the Association for Computational Linguistics and the 11th International Joint Conference on Natural Language Processing (Volume 1: Long Papers)},
  pages={5493--5505},
  year={2021}
}

@inproceedings{zhang2024attentioninstruction,
  title={Can We Instruct LLMs to Compensate for Position Bias?},
  author={Zhang, Meiru and Meng, Zaiqiao and Collier, Nigel},
  booktitle={Findings of the Association for Computational Linguistics: EMNLP 2024},
  pages={12545--12556},
  year={2024}
}

@article{lewis2020retrieval,
  title={Retrieval-augmented generation for knowledge-intensive nlp tasks},
  author={Lewis, Patrick and Perez, Ethan and Piktus, Aleksandra and Petroni, Fabio and Karpukhin, Vladimir and Goyal, Naman and K{"u}ttler, Heinrich and Lewis, Mike and Yih, Wen-tau and Rockt{"a}schel, Tim and others},
  journal={Advances in neural information processing systems},
  volume={33},
  pages={9459--9474},
  year={2020}
}

@article{kamradt2023needle,
  title={{Needle In A Haystack} - Pressure Testing {LLM}s},
   author={Gregory Kamradt},
   year={2023},
   journal ={Github},  
   url={https://github.com/gkamradt/LLMTest_NeedleInAHaystack/tree/main},
}

@inproceedings{yu2024evaluationofrag,
  title={Evaluation of retrieval-augmented generation: A survey},
  author={Yu, Hao and Gan, Aoran and Zhang, Kai and Tong, Shiwei and Liu, Qi and Liu, Zhaofeng},
  booktitle={CCF Conference on Big Data},
  pages={102--120},
  year={2024},
  organization={Springer}
}

@inproceedings{min2019compositional,
  title={Compositional Questions Do Not Necessitate Multi-hop Reasoning},
  author={Min, Sewon and Wallace, Eric and Singh, Sameer and Gardner, Matt and Hajishirzi, Hannaneh and Zettlemoyer, Luke},
  booktitle={Proceedings of the 57th Annual Meeting of the Association for Computational Linguistics},
  pages={4249--4257},
  year={2019}
}

@inproceedings{xiao2024efficientstreaming,
  title={Efficient Streaming Language Models with Attention Sinks},
  author={Xiao, Guangxuan and Tian, Yuandong and Chen, Beidi and Han, Song and Lewis, Mike},
  booktitle={The Twelfth International Conference on Learning Representations},
  year={2024},
}

@article{olsson2022context,
  title={In-context learning and induction heads},
  author={Olsson, Catherine and Elhage, Nelson and Nanda, Neel and Joseph, Nicholas and DasSarma, Nova and Henighan, Tom and Mann, Ben and Askell, Amanda and Bai, Yuntao and Chen, Anna and others},
  journal={arXiv preprint arXiv:2209.11895},
  year={2022}
}

@inproceedings{wang2023primacy,
  title={Primacy Effect of ChatGPT},
  author={Wang, Yiwei and Cai, Yujun and Chen, Muhao and Liang, Yuxuan and Hooi, Bryan},
  booktitle={Proceedings of the 2023 Conference on Empirical Methods in Natural Language Processing},
  pages={108--115},
  year={2023}
}

@inproceedings{yang2018hotpotqa,
  title={HotpotQA: A dataset for diverse, explainable multi-hop question answering},
  author={Yang, Zhilin and Qi, Peng and Zhang, Saizheng and Bengio, Yoshua and Cohen, William and Salakhutdinov, Ruslan and Manning, Christopher D},
  booktitle={Proceedings of the 2018 conference on empirical methods in natural language processing},
  pages={2369--2380},
  year={2018}
}

@inproceedings{an2025whyeffectivelengthfallshort,
  title={Why Does the Effective Context Length of LLMs Fall Short?},
  author={An, Chenxin and Zhang, Jun and Zhong, Ming and Li, Lei and Gong, Shansan and Luo, Yao and Xu, Jingjing and Kong, Lingpeng},
  year = {2025},
  booktitle={The Thirteenth International Conference on Learning Representations}
}

@inproceedings{hsieh2024ruler,
  title={RULER: What's the Real Context Size of Your Long-Context Language Models?},
  author={Hsieh, Cheng-Ping and Sun, Simeng and Kriman, Samuel and Acharya, Shantanu and Rekesh, Dima and Jia, Fei and Ginsburg, Boris},
  booktitle={First Conference on Language Modeling},
  year={2024}
}

@article{yang2025qwen3,
  title={Qwen3 technical report},
  author={Yang, An and Li, Anfeng and Yang, Baosong and Zhang, Beichen and Hui, Binyuan and Zheng, Bo and Yu, Bowen and Gao, Chang and Huang, Chengen and Lv, Chenxu and others},
  journal={arXiv preprint arXiv:2505.09388},
  year={2025}
}

@misc{qwen2025qwen25technicalreport,
      title={Qwen2.5 Technical Report}, 
      author={{Qwen Team} and {Others}},
      year={2025},
      eprint={2412.15115},
      archivePrefix={arXiv},
      primaryClass={cs.CL},
      url={https://arxiv.org/abs/2412.15115}, 
}

@misc{llama3.1_8b_instruct,
  author       = {{Meta Llama Team}},
  title        = {Llama 3.1-8B-Instruct},
  year         = {2024},
  publisher    = {Hugging Face},
  howpublished = {\url{https://huggingface.co/meta-llama/Llama-3.1-8B-Instruct}},
  note         = {Accessed: 2026-01-01}
}

@misc{ministral_8b_instruct_2410,
  author       = {{Mistral AI}},
  title        = {Ministral-8B-Instruct-2410},
  year         = {2024},
  publisher    = {Hugging Face},
  howpublished = {\url{https://huggingface.co/mistralai/Ministral-8B-Instruct-2410}},
  note         = {Accessed: 2026-01-01}
}

@inproceedings{cuconasu2024powerofnoise,
  title={The power of noise: Redefining retrieval for rag systems},
  author={Cuconasu, Florin and Trappolini, Giovanni and Siciliano, Federico and Filice, Simone and Campagnano, Cesare and Maarek, Yoelle and Tonellotto, Nicola and Silvestri, Fabrizio},
  booktitle={Proceedings of the 47th International ACM SIGIR Conference on Research and Development in Information Retrieval},
  pages={719--729},
  year={2024}
}

@inproceedings{press2023measuringandnarrowingcompositionalitygap,
  title={Measuring and narrowing the compositionality gap in language models},
  author={Press, Ofir and Zhang, Muru and Min, Sewon and Schmidt, Ludwig and Smith, Noah A and Lewis, Mike},
  booktitle={Findings of the Association for Computational Linguistics: EMNLP 2023},
  pages={5687--5711},
  year={2023}
}

@article{guo2025deepseekr1,
  title={Deepseek-r1: Incentivizing reasoning capability in llms via reinforcement learning},
  author={Guo, Daya and Yang, Dejian and Zhang, Haowei and Song, Junxiao and Zhang, Ruoyu and Xu, Runxin and Zhu, Qihao and Ma, Shirong and Wang, Peiyi and Bi, Xiao and others},
  journal={arXiv preprint arXiv:2501.12948},
  year={2025}
}

@article{xu2025llmaslogicalreasoner,
  title={Are large language models really good logical reasoners? a comprehensive evaluation and beyond},
  author={Xu, Fangzhi and Lin, Qika and Han, Jiawei and Zhao, Tianzhe and Liu, Jun and Cambria, Erik},
  journal={IEEE Transactions on Knowledge and Data Engineering},
  year={2025},
  publisher={IEEE}
}

@inproceedings{yang2024latentmultihopreasoning,
  title={Do Large Language Models Latently Perform Multi-Hop Reasoning?},
  author={Yang, Sohee and Gribovskaya, Elena and Kassner, Nora and Geva, Mor and Riedel, Sebastian},
  booktitle={Proceedings of the 62nd Annual Meeting of the Association for Computational Linguistics (Volume 1: Long Papers)},
  pages={10210--10229},
  year={2024}
}

@article{schnitzler2024morehopqa,
  title={Morehopqa: More than multi-hop reasoning},
  author={Schnitzler, Julian and Ho, Xanh and Huang, Jiahao and Boudin, Florian and Sugawara, Saku and Aizawa, Akiko},
  journal={arXiv preprint arXiv:2406.13397},
  year={2024}
}

@inproceedings{li2024deceptivesemanticshortcut,
  title={Deceptive Semantic Shortcuts on Reasoning Chains: How Far Can Models Go without Hallucination?},
  author={Li, Bangzheng and Zhou, Ben and Wang, Fei and Fu, Xingyu and Roth, Dan and Chen, Muhao},
  booktitle={Proceedings of the 2024 Conference of the North American Chapter of the Association for Computational Linguistics: Human Language Technologies (Volume 1: Long Papers)},
  pages={7675--7688},
  year={2024}
}

@article{song2025demystifyingdeepsearch,
  title={Demystifying deep search: a holistic evaluation with hint-free multi-hop questions and factorised metrics},
  author={Song, Maojia and Liu, Renhang and Wang, Xinyu and Jiang, Yong and Xie, Pengjun and Huang, Fei and Zhou, Jingren and Herremans, Dorien and Poria, Soujanya},
  journal={arXiv preprint arXiv:2510.05137},
  year={2025}
}

@article{liu2024howmuchraghelpreasoning,
  title={How much can rag help the reasoning of llm?},
  author={Liu, Jingyu and Lin, Jiaen and Liu, Yong},
  journal={arXiv preprint arXiv:2410.02338},
  year={2024}
}

@article{wang2025ragconflictingevidence,
  title={Retrieval-augmented generation with conflicting evidence},
  author={Wang, Han and Prasad, Archiki and Stengel-Eskin, Elias and Bansal, Mohit},
  journal={arXiv preprint arXiv:2504.13079},
  year={2025}
}

@article{wang2025rare,
  title={Rare: Retrieval-augmented reasoning modeling},
  author={Wang, Zhengren and Yu, Jiayang and Ma, Dongsheng and Chen, Zhe and Wang, Yu and Li, Zhiyu and Xiong, Feiyu and Wang, Yanfeng and Tang, Linpeng and Zhang, Wentao and others},
  journal={arXiv preprint arXiv:2503.23513},
  year={2025}
}

@inproceedings{yu2024chainofnote,
  title={Chain-of-note: Enhancing robustness in retrieval-augmented language models},
  author={Yu, Wenhao and Zhang, Hongming and Pan, Xiaoman and Cao, Peixin and Ma, Kaixin and Li, Jian and Wang, Hongwei and Yu, Dong},
  booktitle={Proceedings of the 2024 conference on empirical methods in natural language processing},
  pages={14672--14685},
  year={2024}
}

@inproceedings{li2025structrag,
  title={StructRAG: Boosting Knowledge Intensive Reasoning of LLMs via Inference-time Hybrid Information Structurization},
  author={Li, Zhuoqun and Chen, Xuanang and Yu, Haiyang and Lin, Hongyu and Lu, Yaojie and Tang, Qiaoyu and Huang, Fei and Han, Xianpei and Sun, Le and Li, Yongbin},
  booktitle={The Thirteenth International Conference on Learning Representations},
  year={2025}
}

@inproceedings{adiga2025attentionspeaksvolumes,
  title={Attention speaks volumes: Localizing and mitigating bias in language models},
  author={Adiga, Rishabh and Nushi, Besmira and Chandrasekaran, Varun},
  booktitle={Proceedings of the 63rd Annual Meeting of the Association for Computational Linguistics (Volume 1: Long Papers)},
  pages={26403--26423},
  year={2025}
}

@article{yi2025attentionbasin,
  title={Attention Basin: Why Contextual Position Matters in Large Language Models},
  author={Yi, Zihao and Zeng, Delong and Ling, Zhenqing and Luo, Haohao and Xu, Zhe and Liu, Wei and Luan, Jian and Cao, Wanxia and Shen, Ying},
  journal={arXiv preprint arXiv:2508.05128},
  year={2025}
}

@inproceedings{yu2025mitigateviascalinghiddenstates,
  title={Mitigate Position Bias in LLMs via Scaling a Single Hidden States Channel},
  author={Yu, Yijiong and Jiang, Huiqiang and Luo, Xufang and Wu, Qianhui and Lin, Chin-Yew and Li, Dongsheng and Yang, Yuqing and Huang, Yongfeng and Qiu, Lili},
  booktitle={Findings of the Association for Computational Linguistics: ACL 2025},
  pages={6092--6111},
  year={2025}
}

@article{chen2023extendingcontextwindow,
  title={Extending context window of large language models via positional interpolation},
  author={Chen, Shouyuan and Wong, Sherman and Chen, Liangjian and Tian, Yuandong},
  journal={arXiv preprint arXiv:2306.15595},
  year={2023}
}

@article{ma2025noreasoningcanbeeffective,
  title={Reasoning models can be effective without thinking},
  author={Ma, Wenjie and He, Jingxuan and Snell, Charlie and Griggs, Tyler and Min, Sewon and Zaharia, Matei},
  journal={arXiv preprint arXiv:2504.09858},
  year={2025}
}

@inproceedings{zelikman2024quietstar,
  title={Quiet-STaR: Language Models Can Teach Themselves to Think Before Speaking},
  author={Zelikman, Eric and Harik, Georges and Shao, Yijia and Jayasiri, Varuna and Haber, Nick and Goodman, Noah D},
  booktitle={First Conference on Language Modeling},
  year={2024}
}

@article{weston2023system2attention,
  title={System 2 Attention (is something you might need too)},
  author={Weston, Jason and Sukhbaatar, Sainbayar},
  journal={arXiv preprint arXiv:2311.11829},
  year={2023}
}

@article{li2025system1tosystem2,
  title={From system 1 to system 2: A survey of reasoning large language models},
  author={Li, Zhong-Zhi and Zhang, Duzhen and Zhang, Ming-Liang and Zhang, Jiaxin and Liu, Zengyan and Yao, Yuxuan and Xu, Haotian and Zheng, Junhao and Wang, Pei-Jie and Chen, Xiuyi and others},
  journal={arXiv preprint arXiv:2502.17419},
  year={2025}
}

@article{jaech2024openaio1,
  title={Openai o1 system card},
  author={Jaech, Aaron and Kalai, Adam and Lerer, Adam and Richardson, Adam and El-Kishky, Ahmed and Low, Aiden and Helyar, Alec and Madry, Aleksander and Beutel, Alex and Carney, Alex and others},
  journal={arXiv preprint arXiv:2412.16720},
  year={2024}
}

@inproceedings{kwon2023efficient,
  title={Efficient memory management for large language model serving with pagedattention},
  author={Kwon, Woosuk and Li, Zhuohan and Zhuang, Siyuan and Sheng, Ying and Zheng, Lianmin and Yu, Cody Hao and Gonzalez, Joseph and Zhang, Hao and Stoica, Ion},
  booktitle={Proceedings of the 29th symposium on operating systems principles},
  pages={611--626},
  year={2023}
}

\appendix

\section{Appendix}
\label{sec:appendix}


\subsection{Implementation Details}
\label{appx:implementation}

We utilized vLLM (v0.8.5.post1)~\citep{kwon2023efficient} to perform inference in the default bf16 precision
on A6000, 2x3090, and A100 GPUs. For all experiments,
the temperature was set to 0 to enforce greedy sampling, with the random seed fixed at 42 for reproducibility. Despite greedy decoding, vLLM inference is not strictly bit-deterministic under concurrent batching (due to kernel-level numerical noise and batch-size-dependent attention computations), so exact per-cell EM values may drift by a fraction of a percentage point on re-runs; the directional conclusions we report are robust to this residual stochasticity. Inference on the full dataset required approximately 0.5 hours for standard instruction-following models on 2x3090, faster on A100 and A6000. Qwen3-8B-Think required roughly 3 hours per run. In total, the full Spread and Cross tests for each model involved 150 discrete inference runs.

We selected the tested LLMs based on their maximum context window length and GPU compatibility. This includes Ministral-8B-Instruct, released in 2024, specifically Ministral-8B-Instruct-2410.\footnote{\url{https://huggingface.co/mistralai/Ministral-8B-Instruct-2410}} For Qwen3-8B, we evaluated both ``thinking'' and ``non-thinking'' modes following the official vLLM deployment guidelines.\footnote{\url{https://qwen.readthedocs.io/en/latest/deployment/vllm.html}} 

\subsection{Additional Related Work on System-2 Reasoning}

An emerging paradigm for mitigating reasoning failures involves transitioning from fast System 1 processing to deliberate System-2 reasoning~\citep{li2025system1tosystem2}. Models such as the OpenAI o1 series~\citep{jaech2024openaio1} and DeepSeek-R1~\citep{guo2025deepseekr1} leverage large-scale reinforcement learning to internalize chain-of-thought verification without supervised fine-tuning. Similarly, the Qwen3 series~\citep{yang2025qwen3} unifies ``thinking'' and standard ``non-thinking'' modes within a dynamic compute budget. Beyond explicit reasoning models, techniques like \textit{Quiet-STaR}~\citep{zelikman2024quietstar} and \textit{System-2 Attention}~\citep{weston2023system2attention} act as implicit context reconstruction mechanisms, generating internal rationales or filtering context before attending. Although \citet{ma2025noreasoningcanbeeffective} question the strict necessity of explicit thought tokens in low-budget settings, our work empirically demonstrates that this extended test-time compute is crucial for robustness against position bias. We show that ``thinking'' models actively verify retrieval artifacts, overcoming the topological limitations of standard LLMs.

\subsection{Examples of MuSiQue and NeoQA}
\label{appx:dataset_examples}

\begin{examplebox}{MuSiQue Example 1}
\textbf{Question:} Which county does Lloyd Dane's birthplace belong to?\\
\textbf{Answer:} Miller County\\
\textbf{Gold Doc 1 (meta):} id=para\_3; title=Lloyd Dane; paragraph\_idx=3\\
\textbf{Gold Doc 1 (text):} \textcolor{blue}{Lloyd Dane} (August 19, 1925 – December 11, 2015) was a NASCAR Grand National Series driver from \textcolor{blue}{Eldon, Missouri}. He participated part-time in the 1951 and 1954 to 1964 seasons, capturing four wins, all in his own car...\\
\textbf{Gold Doc 2 (meta):} id=para\_11; title=Eldon, Missouri; paragraph\_idx=11\\
\textbf{Gold Doc 2 (text):} \textcolor{blue}{Eldon} is a city in \textcolor{blue}{Miller County, Missouri}, United States, located thirty miles southwest of Jefferson City...\\
\textbf{Logical Connection:} Lloyd Dane was born in Eldon; Eldon is in Miller County.\\
\textbf{Distractor Example (meta):} id=para\_19; title=Minsk Region\\
\textbf{Distractor Example (text):} Minsk Region or Minsk Voblasć or Minsk Oblast ... is one of the regions of Belarus. Its administrative center is Minsk... 
\end{examplebox}

\begin{examplebox}{MuSiQue Example 2}
\textbf{Question:} Who wrote "Turn Me On" by performer of "Happy Pills"?\\
\textbf{Answer:} John D. Loudermilk\\
\textbf{Gold Doc 1 (meta):} id=para\_0; title=Happy Pills (song); paragraph\_idx=0\\
\textbf{Gold Doc 1 (text):} \textcolor{blue}{"Happy Pills"} is a song by the American singer-songwriter \textcolor{blue}{Norah Jones}. It is the lead single from her fifth studio album "Little Broken Hearts"...\\
\textbf{Gold Doc 2 (meta):} id=para\_10; title=Turn Me On (Mark Dinning song); paragraph\_idx=10\\
\textbf{Gold Doc 2 (text):} \textcolor{blue}{``Turn Me On''} Single by Norah Jones from the album First Sessions ... \textcolor{blue}{Songwriter(s) John D. Loudermilk} ...\\
\textbf{Logical Connection:} Performer of "Happy Pills" is Norah Jones; she performs "Turn Me On"; songwriter is John D. Loudermilk.\\
\textbf{Distractor Example (meta):} id=para\_17; title=Birth control\\
\textbf{Distractor Example (text):} In 1909, Richard Richter developed the first intrauterine device ... Further developments followed in the 1950s... 
\end{examplebox}

\begin{examplebox}{NeoQA Example 1 (time-span question)}
\textbf{Event:} Viroscope app lifecycle and governance changes .\\
\textbf{Question:} What is the total duration between the inferred start of the Viroscope app's six-month testing phase in Misterine City and the release of the report by HealthStream Tech Solutions about the feasibility of the citizen-led data trust model?\\
\textbf{Answer:} 1 year, 2 months, and 28 days\\
\textbf{Gold Doc 1 (meta):} article\_id=epidemics\_3\_0-ev7-3; title=Major Overhaul Planned for Viroscope App...; date=2025-08-12\\
\textbf{Gold Doc 1 (text):} HealthStream Tech Solutions ... released a detailed report ... about adapting \textcolor{blue}{Viroscope} to a citizen-led data trust model. \textcolor{blue}{The report, released on August 12, 2025}, confirms ... a decentralized framework...\\
\textbf{Gold Doc 2 (meta):} article\_id=epidemics\_3\_0-ev0-9; title=Drenvale Institute Launches "Viroscope"...; date=2024-11-15\\
\textbf{Gold Doc 2 (text):} The Drenvale Institute ... unveiled its mobile app \textcolor{blue}{"Viroscope" on November 15, 2024}, in a bold move to revolutionize epidemic tracking ... \textcolor{blue}{"Viroscope" underwent six months of testing in Misterine City}...\\
\textbf{Logical Connection:} Testing lasted 6 months; inferred start is 6 months before 2024-11-15, which is 2024-5-15; compute interval to 2025-08-12, the answer is 1 year, 2 months, and 28 days.\\
\textbf{Distractor Example (meta):} article\_id=epidemics\_3\_0-ev4-10; title=Pilot Study of Viroscope App in Larnwick...\\
\textbf{Distractor Example (text):} The Drenvale Institute for Public Health is moving forward with the pilot rollout of the Viroscope app in Larnwick, scheduled to begin on July 1, 2025...
\end{examplebox}

\begin{examplebox}{NeoQA Example 2}
\textbf{Event:} Norhaven rollout delay and privacy oversight debate in the epidemics timeline (event-centric multi-hop).\\
\textbf{Question:} What specific process confirmed the compliance of the app with data protection standards, as stated by the individual who expressed regret over delays in Norhaven and emphasized public trust?\\
\textbf{Answer:} Independent security audit by a third-party firm\\
\textbf{Gold Doc 1 (meta):} article\_id=epidemics\_3\_0-ev0-11; title="Drenvale Institute Launches 'Viroscope' App Amid Privacy Concerns"; date=2024-11-15\\
\textbf{Gold Doc 1 (text):} The Drenvale Institute for Public Health announced the release of its latest innovation ... addressed privacy concerns around \textcolor{blue}{Viroscope}. Dr. Elara Tovrin ... stated the app does not collect personal identifiers ... \textcolor{blue}{the app underwent an independent security audit conducted by a third-party firm, which confirmed its compliance with data protection standards}...\\
\textbf{Gold Doc 2 (meta):} article\_id=epidemics\_3\_0-ev8-7; title=Privacy Concerns and Oversight Delay Viroscope App Rollout in Norhaven; date=2025-10-05\\
\textbf{Gold Doc 2 (text):} The Viroscope rollout in Norhaven was delayed due to privacy concerns ... \textcolor{blue}{Dr. Elara Tovrin expressed regret over the delay} and \textcolor{blue}{reaffirmed that public trust remains the institute’s top priority}...\\
\textbf{Logical Connection:} Gold Doc 2 identifies the speaker (Dr. Elara Tovrin). Gold Doc 1 states the process she cited as confirming compliance, which is the independent third-party security audit.\\
\textbf{Distractor Example (meta):} article\_id=epidemics\_3\_0-ev8-1; title=Norhaven Launches Review of Viroscope App Amid Privacy Concerns\\
\textbf{Distractor Example (text):} Norhaven’s Ministry of Citizen Protection announced a comprehensive review ... emphasized compliance with the Data Protection Act of 2023 ... announced oversight and evaluation plans...
\end{examplebox}

\subsection{Heatmap Computation and Visualization}
\label{appx:heatmap_details}

We visualize the attention of the first generated answer token, aggregated into semantic spans and differenced across MFAI conditions.

\paragraph{Extraction and aggregation.}
For each instance we render the prompt with the model's chat template and run the forward pass with the \textit{eager} attention implementation, reusing the KV cache so only the target-token attention is computed. Let $a_{l,h,i}$ be that attention weight at layer $l$, head $h$ for input token $i$. The prompt is segmented into $S$ spans: task instruction, attention instruction (when present), question, answer options, and $D$ document blocks. We average over heads to obtain a per-layer span matrix $M_{l,\sigma} = \tfrac{1}{H|\sigma|}\sum_{h,\,i \in \sigma} a_{l,h,i}$, and over a valid-layer range $L_{\text{valid}}=\{1,\dots,26\}$ (excluding layer $0$, which is embedding-dominated, and the last output-specialized layers) to obtain a per-head span matrix $H_{\sigma,h} = \tfrac{1}{|L_{\text{valid}}||\sigma|}\sum_{l \in L_{\text{valid}},\,i \in \sigma} a_{l,h,i}$. We fix one bucket--distance configuration per figure (e.g., $\texttt{gold\_at\_b\_dist1}$ for the Beginning-bucket panel) and average the matrices across $N{=}100$ instances selected from that configuration by a priority-fill sampler: it first takes every Matched-correct / NA-incorrect case (causal-lift cases) and fills the remainder from cases where both modes are correct; Matched-incorrect / NA-correct cases are excluded.

\begin{figure*}[!t]
    \centering
    \begin{minipage}[b]{0.48\textwidth}
        \centering
        \includegraphics[width=\textwidth]{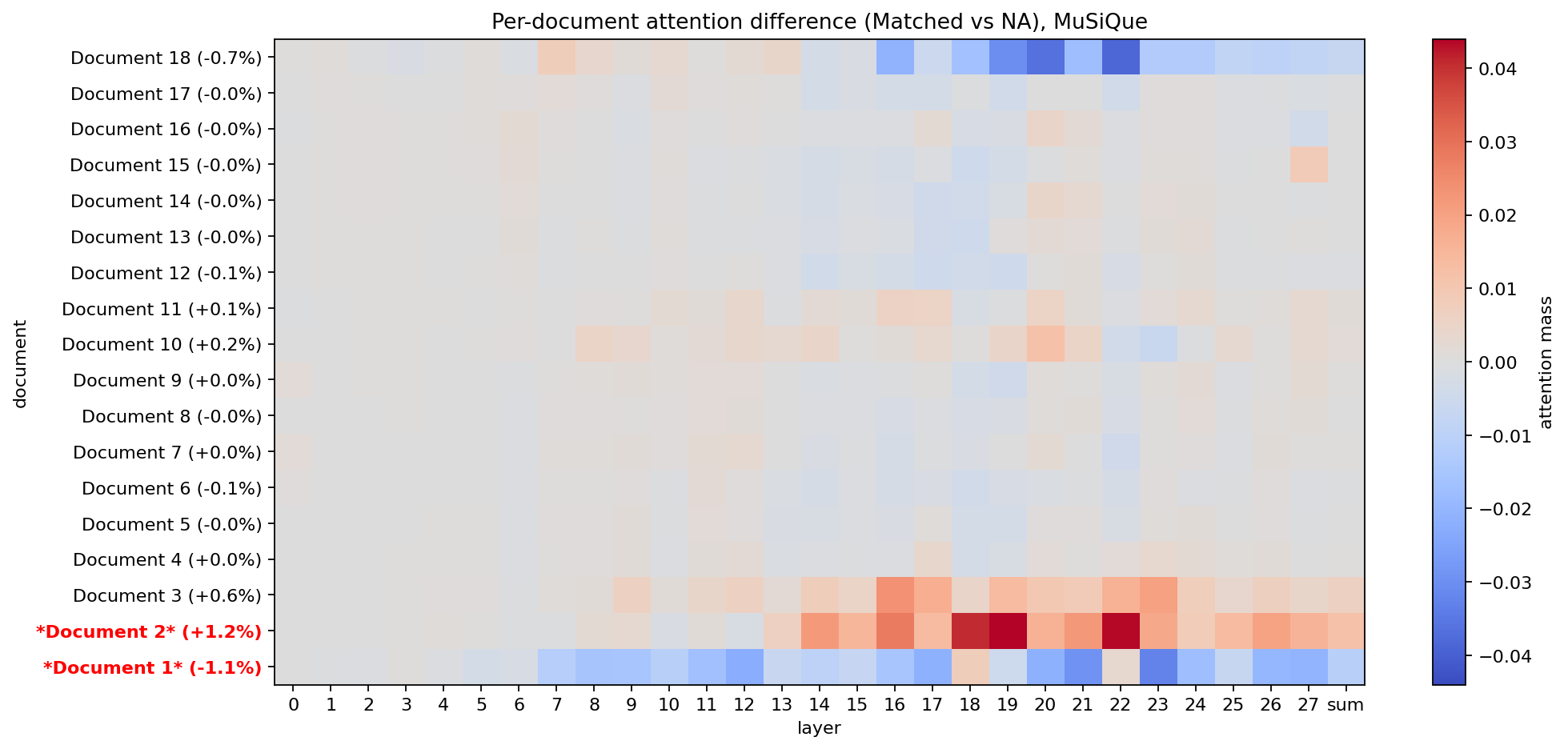}
        \subcaption{Layer-wise Heatmap.}
    \end{minipage}\hfill
    \begin{minipage}[b]{0.48\textwidth}
        \centering
        \includegraphics[width=\textwidth]{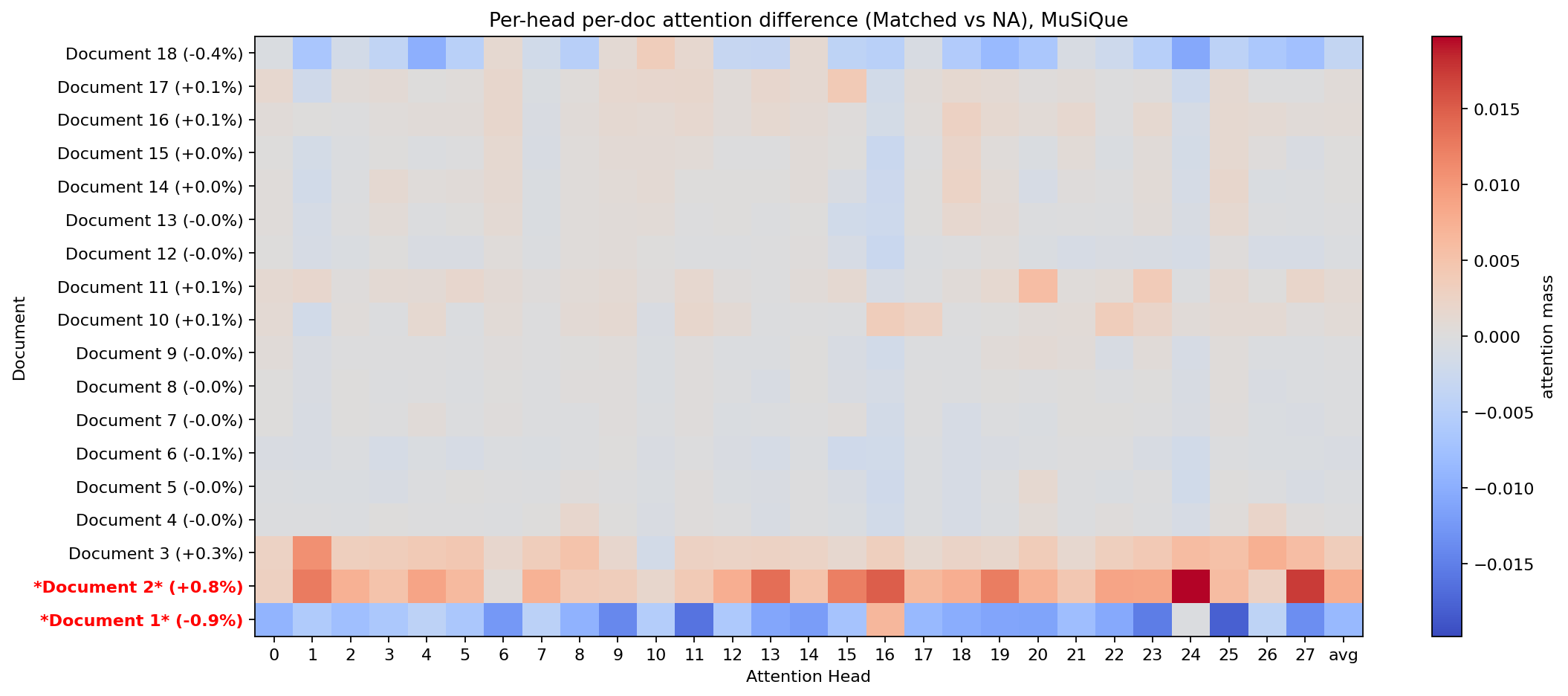}
        \subcaption{Head-wise Heatmap.}
    \end{minipage}
    \caption{Normalized document-share difference (Matched $-$ No MFAI) of Qwen2.5-7B-Instruct on MuSiQue when gold documents are placed in the Beginning bucket (Doc 1 and Doc 2, configuration $\texttt{gold\_at\_b\_dist1}$), averaged over 100 examples drawn with the priority-fill sampler described in Appendix~\ref{appx:heatmap_details}. Y-axis: document spans (with per-document mean over layers/heads, $\times 100$, shown in the label). X-axis: layer index in (a), head index in (b) with a final head-averaged \texttt{avg} column. The colorbar reports the raw signed share difference per cell; e.g., $0.02$ corresponds to a $+2$\,pp reallocation. Red $=$ larger share under Matched; blue $=$ smaller.}
    \label{fig:attn_heatmap_musique}
\end{figure*}

\paragraph{Per-document share difference.}
Figure~\ref{fig:attn_heatmap_musique} shows the quantity we report: the difference, between two modes, of per-layer- (or per-head-) normalized document shares. Restricting $M_{l,\sigma}$ to the $D$ document spans and normalizing so that each layer's doc shares sum to one, $\tilde{M}_{l,d} = M_{l,d} / \sum_{d'} M_{l,d'}$, we plot $\tilde{M}^{\text{matched}}_{l,d} - \tilde{M}^{\text{NA}}_{l,d}$. The per-head version uses the analogous column normalization of $H_{d,h}$, with a final $\texttt{avg}$ column appended that is the row mean over heads. Normalizing before differencing factors out each layer's (or head's) overall doc-attention budget, so colors reflect \emph{reallocation among documents}: red = larger share under Matched, blue = smaller. Gold documents are marked with an asterisk; instruction-targeted documents are highlighted in red. Per-document labels show the mean of the normalized-share difference over layers (or heads), multiplied by $100$ (percentage points); the colorbar reports the raw signed share difference per cell, so a colorbar value of $0.02$ corresponds to a $+2$\,pp reallocation.

\subsection{Statistical Analysis}
\label{appx:statistical-analysis}

We apply two statistical tests to validate the reliability of our experimental results: bootstrap confidence intervals for the accuracy estimates reported in the bar charts, and McNemar's test for the significance of performance differences between the No MFAI and Matched MFAI conditions.

\paragraph{Bootstrap Confidence Intervals.}
Each accuracy value reported in our figures is a point estimate computed over a finite sample of question--answer pairs. To quantify the uncertainty of these estimates, we construct 95\% confidence intervals (CIs) using the bootstrap percentile method~\citep{efron1994introduction}. For a given experimental condition (model, dataset, protocol, position bucket), let $\mathbf{s} = (s_1, \dots, s_N)$ denote the vector of $N$ binary scores. We draw $B = 10{,}000$ bootstrap samples of size $N$ with replacement and compute the replicate accuracy for each. The 95\% CI is given by the 2.5th and 97.5th percentiles of the empirical distribution. These confidence intervals are displayed as error bars on the bar charts in \Cref{fig:spread_test_main,fig:cross_test_main}.

\paragraph{McNemar's Test.}
To test whether Matched MFAI significantly improves accuracy relative to the No MFAI baseline, we apply McNemar's test~\citep{mcnemar1947note}, which is appropriate for comparing paired binary outcomes on the same set of examples. For each setting (model, dataset, protocol, position bucket), we evaluate the same $N$ questions under both conditions, producing paired outcomes $(s_i^{\text{na}}, s_i^{\text{matched}})$ summarized in a $2 \times 2$ contingency table. Let $b$ count questions correct under No MFAI but incorrect under Matched (degraded), and $c$ count the reverse (improved). Under the null hypothesis that Matched MFAI has no effect, the test statistic with continuity correction is:
\begin{equation}
\chi^2 = \frac{(|b - c| - 1)^2}{b + c}
\end{equation}
which follows a $\chi^2$ distribution with one degree of freedom. We report significance at both $p < 0.05$ and $p < 0.01$ thresholds.

\begin{table}[ht]
    \centering
    \small
    \setlength{\tabcolsep}{4pt}
    \resizebox{\columnwidth}{!}{
    \begin{tabular}{llccc}
    \toprule
    \textbf{Dataset} & \textbf{Model} & \textbf{Beginning} & \textbf{Middle} & \textbf{Tail} \\
    \midrule
    \multirow{6}{*}{MuSiQue} & Llama-3.1-8B & $+7.22^{**}$ & $+11.46^{**}$ & $+11.49^{**}$ \\
     & Ministral-8B & $+1.73^{**}$ & $+7.83^{**}$ & $+7.91^{**}$ \\
     & Qwen2.5-7B & $+7.26^{**}$ & $+12.12^{**}$ & $+11.04^{**}$ \\
     & Qwen2.5-14B & $+5.07^{**}$ & $+9.49^{**}$ & $+7.58^{**}$ \\
     & Qwen3-8B & $+2.99^{**}$ & $+8.56^{**}$ & $+8.31^{**}$ \\
     & Qwen3-8B-Think & $+2.57^{**}$ & $+5.26^{**}$ & $+4.83^{**}$ \\
    \midrule
    \multirow{6}{*}{NeoQA} & Llama-3.1-8B & $+2.94^{**}$ & $+4.28^{**}$ & $+2.39^{*}$ \\
     & Ministral-8B & $+6.37^{**}$ & $+3.13^{**}$ & $+4.18^{**}$ \\
     & Qwen2.5-7B & $+3.98^{**}$ & $+0.45$ & $+1.79$ \\
     & Qwen2.5-14B & $+4.23^{**}$ & $+3.33^{**}$ & $+1.14$ \\
     & Qwen3-8B & $+1.79^{*}$ & $+1.14$ & $+0.80$ \\
     & Qwen3-8B-Think & $+2.89^{**}$ & $+2.44^{**}$ & $+0.95$ \\
    \bottomrule
    \end{tabular}
    }
    \caption{McNemar's test: Matched MFAI vs.\ No MFAI on the \textbf{Spread Test} (MuSiQue, NeoQA). Rows are the five core instruction-tuned models plus Qwen3-8B-Think; columns are position buckets. Cell values are $\Delta$ accuracy (\%) = Matched $-$ NA. Significance markers: $^{*}$ $p<0.05$, $^{**}$ $p<0.01$ (McNemar's test with continuity correction; two-sided, paired binary outcomes).}
    \label{tab:mcnemar-spread}
\end{table}

\begin{table}[ht]
    \centering
    \small
    \setlength{\tabcolsep}{4pt}
    \resizebox{\columnwidth}{!}{
    \begin{tabular}{llccc}
    \toprule
    \textbf{Dataset} & \textbf{Model} & \textbf{B+M} & \textbf{B+T} & \textbf{M+T} \\
    \midrule
    \multirow{6}{*}{MuSiQue} & Llama-3.1-8B & $+10.69^{**}$ & $+10.96^{**}$ & $+11.88^{**}$ \\
     & Ministral-8B & $+6.07^{**}$ & $+6.06^{**}$ & $+7.85^{**}$ \\
     & Qwen2.5-7B & $+11.81^{**}$ & $+10.22^{**}$ & $+12.23^{**}$ \\
     & Qwen2.5-14B & $+9.16^{**}$ & $+8.53^{**}$ & $+8.45^{**}$ \\
     & Qwen3-8B & $+6.93^{**}$ & $+7.41^{**}$ & $+8.12^{**}$ \\
     & Qwen3-8B-Think & $+4.24^{**}$ & $+4.79^{**}$ & $+5.35^{**}$ \\
    \midrule
    \multirow{6}{*}{NeoQA} & Llama-3.1-8B & $+3.19^{**}$ & $+2.86^{**}$ & $+1.49$ \\
     & Ministral-8B & $+3.94^{**}$ & $+3.77^{**}$ & $+3.11^{**}$ \\
     & Qwen2.5-7B & $+3.73^{**}$ & $+4.02^{**}$ & $+2.20^{*}$ \\
     & Qwen2.5-14B & $+1.78^{*}$ & $+2.74^{**}$ & $+1.41$ \\
     & Qwen3-8B & $+1.16$ & $+1.99^{*}$ & $+1.78^{*}$ \\
     & Qwen3-8B-Think & $+2.86^{**}$ & $+1.87^{**}$ & $+1.24$ \\
    \bottomrule
    \end{tabular}
    }
    \caption{McNemar's test: Matched MFAI vs.\ No MFAI on the \textbf{Cross Test} (MuSiQue, NeoQA). Rows are the five core instruction-tuned models plus Qwen3-8B-Think; columns are bucket pairs (Beginning+Middle, Beginning+Tail, Middle+Tail). Cell values are $\Delta$ accuracy (\%) = Matched $-$ NA. Significance markers: $^{*}$ $p<0.05$, $^{**}$ $p<0.01$ (McNemar's test with continuity correction; two-sided, paired binary outcomes).}
    \label{tab:mcnemar-cross}
\end{table}

\Cref{tab:mcnemar-spread,tab:mcnemar-cross} report the paired McNemar results for the five core instruction-tuned models plus Qwen3-8B-Think on MuSiQue and NeoQA, under the Spread and Cross protocols respectively. On MuSiQue, Matched MFAI significantly improves over No MFAI in every cell at $p<0.01$ (36/36 across both protocols), consistent with the recognition-bottleneck interpretation: explicitly steering attention toward gold documents reliably rescues performance. On NeoQA, the effect is smaller and the significance pattern is mixed: most cells still reach $p<0.01$ or $p<0.05$, but a few cells with small $\Delta$ (e.g., Qwen2.5-7B Spread Middle/Tail, Qwen3-8B Cross B+M) do not reach significance, reflecting the smaller effect sizes on NeoQA's more homogeneous same-timeline context (No-MFAI baseline accuracies of $\sim$50--70\% on NeoQA vs.\ $\sim$10--35\% on MuSiQue). Per protocol, the Spread test yields 28/36 cells at $p<0.01$ and 30/36 at $p<0.05$, while the Cross test yields 28/36 at $p<0.01$ and 32/36 at $p<0.05$; aggregated, $56/72$ cells reach $p<0.01$ and $62/72$ reach $p<0.05$.

\subsection{Ablation}
\label{appx:ablation}

\paragraph{Gold-Only Ablation.}
To establish an upper bound on model performance and isolate the impact of distractors, we conducted a gold-only ablation where models receive only the two gold documents. \Cref{tab:gold_only_ablation} presents the results. This setting represents the ideal scenario where recognition is perfect and the model only needs to perform reasoning or synthesis over the minimal necessary context. The performance gap between this gold-only condition and the full 18-document setting (reported in \Cref{fig:spread_test_main} in \Cref{sec:bias-distance}) quantifies the cost of distraction, revealing the extent to which noise filtering bottlenecks each model. 

\begin{table}[ht]
    \centering
    \small
    \setlength{\tabcolsep}{4pt}
    \resizebox{\columnwidth}{!}{
    \begin{tabular}{lccccc}
    \toprule
    \textbf{Model} & \textbf{NeoQA} & \multicolumn{3}{c}{\textbf{MuSiQue EM (\%)}} \\
    \cmidrule(lr){3-5}
    & Acc.\ (\%) & 2h & 3h & 4h \\
    \midrule
    Llama-3.1-8B-Instruct & 66.42 & 40.45 & 35.54 & 32.84 \\
    Ministral-8B-Instruct & 63.18 & 37.40 & 25.76 & 34.81 \\
    Qwen2.5-7B-Instruct & 68.91 & 19.58 & 15.19 & 17.78 \\
    Qwen2.5-14B-Instruct & 69.15 & 41.57 & 37.65 & 48.15 \\
    Qwen3-8B & 71.64 & 42.46 & 34.21 & 42.22 \\
    Qwen3-8B-Think & 71.64 & 44.70 & --- & --- \\
    \bottomrule
    \end{tabular}
    }
    \caption{Gold-only exact-match accuracy (\%) for NeoQA and MuSiQue at 2-hop, 3-hop, and 4-hop. Each setting feeds the model only the gold supporting documents (no distractors). ``---'' marks cells where the experiment was not run for that model. Qwen3-8B-Think has 2-hop only.}
    \label{tab:gold_only_ablation}
\end{table}

Specifically, all models, including Qwen3-8B-Think, exhibit performance degradation when distractors are present compared to the high-visibility Beginning bucket on MuSiQue. However, this degradation is typically less severe than the drop caused by position bias. For example, the performance of Ministral-8B-Instruct drops by 7.69\% due to distractors when gold documents are in the Beginning bucket, but moving the gold documents to the Middle bucket leads to 11.04\% drop. Performance on NeoQA also declines when distracting documents are added, though at a similar level to the drop caused by position bias. Surprisingly, Qwen3-8B-Think appears to benefit from the noise, which suggests that distractors may trigger more rigorous verification in ``thinking'' models. The gold-only upper bound generally drops from 2-hop to 3-hop (see 3-hop and 4-hop columns of \Cref{tab:gold_only_ablation}), though the trend is non-monotone at 4-hop for several models.

\paragraph{No-Document Ablation.}
To verify that model performance is driven by the provided context rather than parametric memory, we removed all documents from the input. As shown in \Cref{tab:no_document_ablation}, accuracy drops to near-zero across all models for MuSiQue at all hop counts (2-, 3-, and 4-hop) and for NeoQA, with most models correctly outputting ``Unanswerable'' (except Qwen2.5-7B-Instruct). This confirms that models are indeed relying on the retrieved documents rather than memorized knowledge, validating the integrity of our experimental setup even as hop count increases.

\begin{table}[ht]
    \centering
    \small
    \setlength{\tabcolsep}{4pt}
    \resizebox{\columnwidth}{!}{
    \begin{tabular}{lccccc}
    \toprule
    \textbf{Model} & \multicolumn{2}{c}{\textbf{NeoQA}} & \multicolumn{3}{c}{\textbf{MuSiQue EM (\%)}} \\
    \cmidrule(lr){2-3} \cmidrule(lr){4-6}
    & Acc.\ (\%) & Unans.\ (\%) & 2h & 3h & 4h \\
    \midrule
    Llama-3.1-8B-Instruct &  0.00 & 100.00 &  0.08 &  0.26 &  0.25 \\
    Ministral-8B-Instruct &  0.00 & 100.00 &  1.77 &  0.00 &  0.00 \\
    Qwen2.5-7B-Instruct & 14.93 & 61.44 &  0.00 &  0.00 &  0.00 \\
    Qwen2.5-14B-Instruct &  0.00 & 100.00 &  0.00 &  0.26 &  0.25 \\
    Qwen3-8B &  1.74 & 94.03 &  0.16 &  0.00 &  0.25 \\
    Qwen3-8B-Think &  0.00 & 100.00 &  0.08 & --- & --- \\
    \bottomrule
    \end{tabular}
    }
    \caption{No-document ablation exact-match accuracy (\%) for NeoQA and MuSiQue at 2-hop, 3-hop, and 4-hop. Documents are removed from the prompt; NeoQA ``Unans.'' is the rate at which the model selects ``Unanswerable''. ``---'' marks cells where the experiment was not run.}
    \label{tab:no_document_ablation}
\end{table}

\subsection{Supplementary Validation: Additional Datasets and Model Scale}
\label{appx:supplementary-validation}

To assess the generalizability of the Weakest Link Effect beyond the two primary datasets and the five core models, we conducted supplementary evaluations on additional datasets, hop counts, and a larger model.

\subsubsection{2WikiMultiHopQA}

We evaluated the Compositional and Inference subsets of 2WikiMultiHopQA (2-hop; 2{,}685 and 736 examples respectively) on the five core models. The key patterns from MuSiQue replicate across both subsets. To reduce compute load on this supplementary evaluation, we sampled a sparse subset of the inter-gold distance and cross local-index grid used for MuSiQue and NeoQA: Spread distances $d \in \{1, 3, 5\}$ (vs.\ $1$--$5$ on MuSiQue/NeoQA) and Cross local indices $\in \{0, 3, 5\}$ (vs.\ $0$--$5$), which still spans the within-bucket range.\footnote{Within-bucket variation is small on MuSiQue (cf.\ \Cref{fig:spread_test_main}); sampling every other distance and index was chosen to preserve the near-extreme cases while halving GPU cost.}

\paragraph{Step-function position bias.} \Cref{tab:2wiki-spread-distance} reports the Spread NA accuracy per bucket across inter-gold distances 1, 3, and 5. Within-bucket variation is generally small (particularly in the Compositional subset), but shifting the evidence set between buckets causes a large drop, replicating the step-function pattern observed on MuSiQue.

\begin{table}[ht]
    \centering
    \small
    \setlength{\tabcolsep}{4pt}
    \resizebox{\columnwidth}{!}{
    \begin{tabular}{llccccc}
    \toprule
    \textbf{Subset} & \textbf{Model} & \textbf{Bucket} & \textbf{$d=1$} & \textbf{$d=3$} & \textbf{$d=5$} & \textbf{Avg} \\
    \midrule
    \multirow{15}{*}{Comp.}
        & \multirow{3}{*}{Llama-3.1-8B} & Beginning & 35.20 & 34.67 & 33.82 & 34.56 \\
        &                               & Middle & 24.88 & 24.99 & 24.62 & 24.83 \\
        &                               & Tail & 23.02 & 21.86 & 25.03 & 23.30 \\
        & \multirow{3}{*}{Ministral-8B} & Beginning & 41.08 & 39.74 & 38.14 & 39.65 \\
        &                               & Middle & 29.91 & 28.90 & 27.86 & 28.89 \\
        &                               & Tail & 25.74 & 24.51 & 24.80 & 25.02 \\
        & \multirow{3}{*}{Qwen2.5-7B} & Beginning & 28.23 & 26.52 & 27.00 & 27.25 \\
        &                               & Middle & 16.95 & 16.09 & 15.98 & 16.34 \\
        &                               & Tail & 16.57 & 16.65 & 19.44 & 17.55 \\
        & \multirow{3}{*}{Qwen2.5-14B} & Beginning & 43.17 & 40.97 & 40.60 & 41.58 \\
        &                               & Middle & 35.83 & 34.53 & 34.30 & 34.89 \\
        &                               & Tail & 35.08 & 34.64 & 34.71 & 34.81 \\
        & \multirow{3}{*}{Qwen3-8B} & Beginning & 41.12 & 39.93 & 38.47 & 39.84 \\
        &                               & Middle & 31.62 & 31.40 & 29.98 & 31.00 \\
        &                               & Tail & 29.46 & 29.05 & 29.68 & 29.40 \\
    \midrule
    \multirow{15}{*}{Infer.}
        & \multirow{3}{*}{Llama-3.1-8B} & Beginning & 30.98 & 28.80 & 25.68 & 28.49 \\
        &                               & Middle & 13.86 & 12.77 & 12.09 & 12.91 \\
        &                               & Tail & 12.64 & 10.87 & 12.91 & 12.14 \\
        & \multirow{3}{*}{Ministral-8B} & Beginning & 31.66 & 22.83 & 20.65 & 25.05 \\
        &                               & Middle &  9.24 &  8.56 &  8.70 &  8.83 \\
        &                               & Tail &  7.34 &  6.25 &  7.74 &  7.11 \\
        & \multirow{3}{*}{Qwen2.5-7B} & Beginning &  7.88 &  7.07 &  7.74 &  7.56 \\
        &                               & Middle &  5.43 &  4.76 &  4.89 &  5.03 \\
        &                               & Tail &  5.98 &  4.62 &  6.11 &  5.57 \\
        & \multirow{3}{*}{Qwen2.5-14B} & Beginning & 36.55 & 28.53 & 25.00 & 30.03 \\
        &                               & Middle & 13.04 & 12.64 & 10.19 & 11.96 \\
        &                               & Tail & 10.87 &  9.51 & 10.46 & 10.28 \\
        & \multirow{3}{*}{Qwen3-8B} & Beginning & 23.51 & 19.02 & 17.53 & 20.02 \\
        &                               & Middle &  8.56 &  8.15 &  7.61 &  8.11 \\
        &                               & Tail &  6.25 &  6.25 &  6.11 &  6.20 \\
    \bottomrule
    \end{tabular}
    }
    \caption{2WikiMultiHopQA Spread NA accuracy (\%) per bucket across inter-gold distances 1, 3, and 5.}
    \label{tab:2wiki-spread-distance}
\end{table}

\paragraph{MFAI effects on Spread.} \Cref{tab:2wiki-spread-mfai} shows that Matched MFAI consistently rescues the low-visibility buckets (positive $\Delta$), while Unmatched MFAI degrades performance (negative $\Delta$), demonstrating the asymmetric response to correct vs.\ misleading attention steering.

\begin{table}[ht]
    \centering
    \small
    \setlength{\tabcolsep}{3pt}
    \resizebox{\columnwidth}{!}{
    \begin{tabular}{llcccc}
    \toprule
    \textbf{Subset} & \textbf{Model} & \textbf{Bucket} & \textbf{NA} & \textbf{Matched ($\Delta$)} & \textbf{Unmatched ($\Delta$)} \\
    \midrule
    \multirow{15}{*}{Comp.}
        & \multirow{3}{*}{Llama-3.1-8B} & Beginning & 34.56 & 38.83 (+4.27) & 32.45 (-2.12) \\
        &                               & Middle & 24.83 & 37.27 (+12.44) & 23.34 (-1.49) \\
        &                               & Tail & 23.30 & 36.82 (+13.52) & 21.92 (-1.38) \\
        & \multirow{3}{*}{Ministral-8B} & Beginning & 39.65 & 40.09 (+0.43) & 38.04 (-1.61) \\
        &                               & Middle & 28.89 & 33.53 (+4.64) & 26.44 (-2.45) \\
        &                               & Tail & 25.02 & 30.75 (+5.74) & 21.99 (-3.03) \\
        & \multirow{3}{*}{Qwen2.5-7B} & Beginning & 27.25 & 32.97 (+5.72) & 24.54 (-2.71) \\
        &                               & Middle & 16.34 & 28.38 (+12.04) & 13.73 (-2.61) \\
        &                               & Tail & 17.55 & 28.23 (+10.68) & 15.56 (-2.00) \\
        & \multirow{3}{*}{Qwen2.5-14B} & Beginning & 41.58 & 43.70 (+2.12) & 28.00 (-13.58) \\
        &                               & Middle & 34.89 & 42.83 (+7.95) & 26.18 (-8.71) \\
        &                               & Tail & 34.81 & 41.01 (+6.19) & 29.26 (-5.56) \\
        & \multirow{3}{*}{Qwen3-8B} & Beginning & 39.84 & 39.86 (+0.02) & 34.58 (-5.26) \\
        &                               & Middle & 31.00 & 37.01 (+6.01) & 26.36 (-4.64) \\
        &                               & Tail & 29.40 & 35.12 (+5.72) & 24.63 (-4.77) \\
    \midrule
    \multirow{15}{*}{Infer.}
        & \multirow{3}{*}{Llama-3.1-8B} & Beginning & 28.49 & 37.14 (+8.65) & 22.42 (-6.07) \\
        &                               & Middle & 12.91 & 26.95 (+14.04) & 12.79 (-0.11) \\
        &                               & Tail & 12.14 & 24.05 (+11.91) & 12.70 (+0.57) \\
        & \multirow{3}{*}{Ministral-8B} & Beginning & 25.05 & 28.58 (+3.53) & 21.63 (-3.42) \\
        &                               & Middle &  8.83 & 15.44 (+6.61) &  7.93 (-0.91) \\
        &                               & Tail &  7.11 & 13.09 (+5.98) &  6.16 (-0.95) \\
        & \multirow{3}{*}{Qwen2.5-7B} & Beginning &  7.56 & 15.44 (+7.88) &  7.16 (-0.41) \\
        &                               & Middle &  5.03 & 12.86 (+7.84) &  4.82 (-0.20) \\
        &                               & Tail &  5.57 & 11.73 (+6.16) &  5.71 (+0.14) \\
        & \multirow{3}{*}{Qwen2.5-14B} & Beginning & 30.03 & 33.42 (+3.40) & 17.57 (-12.45) \\
        &                               & Middle & 11.96 & 21.51 (+9.56) &  8.88 (-3.08) \\
        &                               & Tail & 10.28 & 19.34 (+9.06) &  8.74 (-1.54) \\
        & \multirow{3}{*}{Qwen3-8B} & Beginning & 20.02 & 25.27 (+5.25) & 13.99 (-6.02) \\
        &                               & Middle &  8.11 & 17.98 (+9.87) &  5.82 (-2.29) \\
        &                               & Tail &  6.20 & 15.35 (+9.15) &  5.41 (-0.79) \\
    \bottomrule
    \end{tabular}
    }
    \caption{2WikiMultiHopQA Spread protocol MFAI effects per bucket. $\Delta$ is relative to NA in the same cell.}
    \label{tab:2wiki-spread-mfai}
\end{table}

\begin{table}[ht]
    \centering
    \small
    \setlength{\tabcolsep}{4pt}
    \resizebox{\columnwidth}{!}{
    \begin{tabular}{llccccc}
    \toprule
    \textbf{Subset} & \textbf{Model} & \textbf{Pair} & \textbf{idx 0} & \textbf{idx 3} & \textbf{idx 5} & \textbf{Avg} \\
    \midrule
    \multirow{15}{*}{Comp.}
        & \multirow{3}{*}{Llama-3.1-8B} & B+M & 33.67 & 27.15 & 24.58 & 28.47 \\
        &                               & B+T & 32.51 & 26.07 & 26.37 & 28.32 \\
        &                               & M+T & 24.88 & 22.57 & 24.92 & 24.12 \\
        & \multirow{3}{*}{Ministral-8B} & B+M & 37.09 & 29.68 & 28.12 & 31.63 \\
        &                               & B+T & 35.64 & 26.85 & 27.52 & 30.01 \\
        &                               & M+T & 26.82 & 25.14 & 25.07 & 25.67 \\
        & \multirow{3}{*}{Qwen2.5-7B} & B+M & 25.92 & 17.13 & 15.98 & 19.68 \\
        &                               & B+T & 25.70 & 17.84 & 20.37 & 21.30 \\
        &                               & M+T & 15.72 & 17.54 & 19.52 & 17.59 \\
        & \multirow{3}{*}{Qwen2.5-14B} & B+M & 40.37 & 35.83 & 34.64 & 36.95 \\
        &                               & B+T & 39.96 & 35.31 & 35.27 & 36.85 \\
        &                               & M+T & 33.93 & 34.08 & 34.82 & 34.28 \\
        & \multirow{3}{*}{Qwen3-8B} & B+M & 37.73 & 32.81 & 30.61 & 33.72 \\
        &                               & B+T & 36.31 & 30.91 & 30.50 & 32.58 \\
        &                               & M+T & 29.80 & 28.60 & 29.16 & 29.19 \\
    \midrule
    \multirow{15}{*}{Infer.}
        & \multirow{3}{*}{Llama-3.1-8B} & B+M & 24.18 & 14.81 & 12.77 & 17.26 \\
        &                               & B+T & 22.15 & 14.27 & 13.45 & 16.62 \\
        &                               & M+T & 12.64 & 10.33 & 11.41 & 11.46 \\
        & \multirow{3}{*}{Ministral-8B} & B+M & 19.16 & 10.73 &  9.24 & 13.04 \\
        &                               & B+T & 16.30 &  9.51 &  8.97 & 11.59 \\
        &                               & M+T &  8.56 &  7.07 &  7.34 &  7.65 \\
        & \multirow{3}{*}{Qwen2.5-7B} & B+M &  7.61 &  3.80 &  5.57 &  5.66 \\
        &                               & B+T &  7.74 &  3.80 &  5.71 &  5.75 \\
        &                               & M+T &  4.89 &  4.76 &  5.84 &  5.16 \\
        & \multirow{3}{*}{Qwen2.5-14B} & B+M & 25.54 & 13.32 & 11.55 & 16.80 \\
        &                               & B+T & 20.65 & 13.18 & 11.82 & 15.22 \\
        &                               & M+T & 11.28 & 10.19 & 10.05 & 10.51 \\
        & \multirow{3}{*}{Qwen3-8B} & B+M & 17.66 &  8.97 &  7.34 & 11.32 \\
        &                               & B+T & 14.54 &  8.70 &  8.42 & 10.55 \\
        &                               & M+T &  7.61 &  6.79 &  6.39 &  6.93 \\
    \bottomrule
    \end{tabular}
    }
    \caption{2WikiMultiHopQA Cross NA accuracy (\%) at local indices 0, 3, and 5 within each bucket-pair.}
    \label{tab:2wiki-cross-local-idx}
\end{table}

\paragraph{Cross local-index invariance.} \Cref{tab:2wiki-cross-local-idx} shows that Cross accuracy is relatively stable across local indices (0, 3, 5) within each bucket-pair, indicating that bucket-pair membership rather than fine-grained offset is the primary driver of performance.

\paragraph{Weakest Link Effect on Cross.} \Cref{tab:2wiki-cross-weakest-link} reports the Cross protocol results with Weakest Link references. Cross NA typically falls between Spread Min and Spread Avg and sits well below the stronger constituent bucket, confirming the Weakest Link Effect on 2WikiMultiHopQA.

\subsubsection{MuSiQue 3-Hop and 4-Hop}
\begin{table}[ht]
    \centering
    \small
    \setlength{\tabcolsep}{6pt}
    \begin{tabular}{llccc}
    \toprule
    \textbf{Model} & \textbf{Hop} & \textbf{All-B} & \textbf{All-M} & \textbf{All-T} \\
    \midrule
    \multirow{2}{*}{Llama-3.1-8B} & 3h & 25.10 & 19.55 & 19.68 \\
                                   & 4h & 21.48 & 15.31 & 16.54 \\
    \multirow{2}{*}{Ministral-8B} & 3h & 20.87 & 13.47 &  9.64 \\
                                   & 4h & 19.01 & 16.30 & 15.56 \\
    \multirow{2}{*}{Qwen2.5-7B} & 3h & 11.10 & 12.02 & 12.55 \\
                                   & 4h & 16.54 & 14.32 & 12.35 \\
    \multirow{2}{*}{Qwen2.5-14B} & 3h & 32.36 & 23.65 & 22.85 \\
                                   & 4h & 33.58 & 27.41 & 27.65 \\
    \multirow{2}{*}{Qwen3-8B} & 3h & 29.85 & 23.25 & 20.34 \\
                                   & 4h & 39.26 & 30.62 & 28.64 \\
    \bottomrule
    \end{tabular}
    \caption{MuSiQue multi-hop single-bucket baseline accuracy (\%) at 3-hop and 4-hop, with all gold documents placed entirely in the Beginning / Middle / Tail bucket.}
    \label{tab:multihop-single-bucket}
\end{table}

The MuSiQue 3-hop (757 examples) and 4-hop (405 examples) subsets confirm that the Weakest Link Effect generalizes to longer reasoning chains.

\paragraph{Placement configurations.} Each placement label concatenates \textit{bucket-letter + local-index} tokens per gold document, with bucket letters B (global $0$--$5$), M ($6$--$11$), T ($12$--$17$) and local index $\in\{0,\dots,5\}$. For example, \texttt{b0b1b2} places three golds in Beginning; \texttt{b0b1m2} is a ``2B+1M'' split; 4-hop labels extend the pattern (\texttt{b0b1b2b3}, \texttt{b0b1b2m3}, etc.). \Cref{tab:multihop-single-bucket} uses the all-in-one-bucket configs (\texttt{b0b1b2}, \texttt{m0m1m2}, \texttt{t0t1t2} and 4-hop analogues). \Cref{tab:multihop-cross-weakest-link} uses the mixed configs \texttt{2B+1M} (\texttt{b0b1m2} / \texttt{b0b1b2m3}) and \texttt{2B+1T} (\texttt{b0b1t2} / \texttt{b0b1b2t3}). \textit{Spread Max} is the all-Beginning reference; \textit{Spread Min} is the all-one-bucket NA for the weaker constituent (all-M or all-T).

\paragraph{Unmatched mirror.} The \textit{Unmatched} column in \Cref{tab:multihop-cross-weakest-link} keeps the gold documents in place but redirects the MFAI instruction into an empty (gold-free) bucket at the \emph{same local indices} as the golds --- i.e., the local-index pattern is preserved, only the bucket letter is swapped. When more than one bucket is empty, we choose by priority T $>$ M $>$ B. Concretely: \texttt{b0b1m2} (B, M occupied) $\to$ \texttt{t0t1t2}; \texttt{b0b1b2} (only B occupied) $\to$ \texttt{t0t1t2} (T preferred over M); \texttt{b0b1b2t3} (B, T occupied) $\to$ \texttt{m0m1m2m3}. Configurations that fill all three buckets (e.g., \texttt{b0m1t2}) are excluded because no empty bucket is available for the mirror.

\paragraph{Single-bucket position bias.} \Cref{tab:multihop-single-bucket} reports accuracy when all gold documents are placed in a single bucket. Position bias persists at higher hop counts: Beginning yields substantially higher accuracy than Middle or Tail for most models.

\paragraph{Cross Weakest Link.} \Cref{tab:multihop-cross-weakest-link} shows that when gold documents are split between buckets, Cross performance approaches the weaker single-bucket reference (Spread Min) rather than the average, and Matched MFAI rescues the drop --- confirming the Weakest Link Effect at 3-hop and 4-hop.

\subsubsection{Model Scale: Qwen2.5-32B}

To address whether the Weakest Link Effect persists at larger model scales, we evaluated Qwen2.5-32B-Instruct-GPTQ-Int8 on a subset comprising the first 600 MuSiQue examples. The findings align with the smaller models along three axes.

\paragraph{Step-function at 32B.} \Cref{tab:32b-spread-distance} shows that within-bucket distance variation remains small while between-bucket variation is substantial, replicating the MuSiQue pattern at larger scale.

\begin{table}[ht]
    \centering
    \small
    \setlength{\tabcolsep}{6pt}
    \begin{tabular}{lcccc}
    \toprule
    \textbf{Bucket} & \textbf{$d=1$} & \textbf{$d=3$} & \textbf{$d=5$} & \textbf{Avg} \\
    \midrule
    Beginning   & 40.33 & 38.00 & 36.33 & 38.22 \\
    Middle      & 32.83 & 32.17 & 31.00 & 32.00 \\
    Tail        & 32.67 & 31.83 & 33.83 & 32.78 \\
    \bottomrule
    \end{tabular}
    \caption{Qwen2.5-32B-Instruct-GPTQ-Int8 Spread NA accuracy (\%) per bucket across inter-gold distances 1, 3, and 5 (first 600 MuSiQue examples).}
    \label{tab:32b-spread-distance}
\end{table}

\paragraph{MFAI at 32B.} \Cref{tab:32b-spread-mfai} confirms that Matched MFAI rescues low-visibility buckets and Unmatched degrades performance, mirroring the smaller-model pattern.

\begin{table}[ht]
    \centering
    \small
    \setlength{\tabcolsep}{6pt}
    \begin{tabular}{lccc}
    \toprule
    \textbf{Bucket} & \textbf{NA} & \textbf{Matched ($\Delta$)} & \textbf{Unmatched ($\Delta$)} \\
    \midrule
    Beginning   & 38.22 & 47.06 (+8.83) & 19.00 (-19.22) \\
    Middle      & 32.00 & 44.39 (+12.39) & 21.11 (-10.89) \\
    Tail        & 32.78 & 44.72 (+11.94) & 28.72 (-4.06) \\
    \bottomrule
    \end{tabular}
    \caption{Qwen2.5-32B-Instruct-GPTQ-Int8 Spread protocol MFAI effects per bucket, pooled across distances 1, 3, 5. $\Delta$ is relative to NA.}
    \label{tab:32b-spread-mfai}
\end{table}

\paragraph{Weakest Link at 32B.} \Cref{tab:32b-cross-weakest-link} shows that Cross performance falls at or below Spread Min and is fully restored by Matched MFAI, confirming the recognition bottleneck persists at scale.

\begin{table*}[!htbp]
    \centering
    \small
    \setlength{\tabcolsep}{4pt}
    \begin{tabular}{llccccccc}
    \toprule
    \textbf{Model} & \textbf{Config} & \textbf{Cross NA} & \textbf{Cross Matched} & \textbf{Cross Unmatched} & \textbf{Spread Min} & \textbf{Spread Max} & \textbf{$\Delta$ Max} \\
    \midrule
    \multirow{4}{*}{Llama-3.1-8B} & 3h 2B+1M & 21.00 & 32.63 & 17.97 & 19.55 & 25.10 & $-$ 4.10 \\
                                   & 3h 2B+1T & 19.42 & 30.52 & 17.44 & 19.68 & 25.10 & $-$ 5.68 \\
                                   & 4h 3B+1M & 18.27 & 31.11 & 20.49 & 15.31 & 21.48 & $-$ 3.21 \\
                                   & 4h 3B+1T & 15.80 & 29.38 & 17.04 & 16.54 & 21.48 & $-$ 5.68 \\
    \multirow{4}{*}{Ministral-8B} & 3h 2B+1M & 14.13 & 16.51 & 12.68 & 13.47 & 20.87 & $-$ 6.74 \\
                                   & 3h 2B+1T &  9.64 & 13.87 &  8.45 &  9.64 & 20.87 & $-$11.23 \\
                                   & 4h 3B+1M & 14.81 & 15.56 & 14.07 & 16.30 & 19.01 & $-$ 4.20 \\
                                   & 4h 3B+1T & 14.57 & 15.56 & 13.83 & 15.56 & 19.01 & $-$ 4.44 \\
    \multirow{4}{*}{Qwen2.5-7B} & 3h 2B+1M &  8.72 & 18.23 &  5.81 & 12.02 & 11.10 & $-$ 2.38 \\
                                   & 3h 2B+1T &  8.98 & 16.78 &  7.27 & 12.55 & 11.10 & $-$ 2.11 \\
                                   & 4h 3B+1M & 13.58 & 22.96 & 10.62 & 14.32 & 16.54 & $-$ 2.96 \\
                                   & 4h 3B+1T & 13.33 & 21.23 & 12.59 & 12.35 & 16.54 & $-$ 3.21 \\
    \multirow{4}{*}{Qwen2.5-14B} & 3h 2B+1M & 26.29 & 33.03 & 15.72 & 23.65 & 32.36 & $-$ 6.08 \\
                                   & 3h 2B+1T & 23.51 & 28.27 & 14.93 & 22.85 & 32.36 & $-$ 8.85 \\
                                   & 4h 3B+1M & 31.60 & 39.75 & 20.74 & 27.41 & 33.58 & $-$ 1.98 \\
                                   & 4h 3B+1T & 30.12 & 36.54 & 18.77 & 27.65 & 33.58 & $-$ 3.46 \\
    \multirow{4}{*}{Qwen3-8B} & 3h 2B+1M & 24.17 & 28.93 & 16.51 & 23.25 & 29.85 & $-$ 5.68 \\
                                   & 3h 2B+1T & 21.66 & 28.01 & 15.06 & 20.34 & 29.85 & $-$ 8.19 \\
                                   & 4h 3B+1M & 32.59 & 38.52 & 27.90 & 30.62 & 39.26 & $-$ 6.67 \\
                                   & 4h 3B+1T & 30.12 & 34.57 & 24.69 & 28.64 & 39.26 & $-$ 9.14 \\
    \bottomrule
    \end{tabular}
    \caption{MuSiQue multi-hop Cross Weakest Link accuracy (\%) for mixed-bucket gold placements. Config denotes gold-document distribution (e.g., 2B+1M = 2 in Beginning, 1 in Middle). Cross NA / Matched / Unmatched come from the mixed-placement files; Spread Min is the weaker single-bucket reference (M or T), Spread Max is the all-B single-bucket reference. $\Delta$ Max = Cross NA $-$ Spread Max.}
    \label{tab:multihop-cross-weakest-link}
\end{table*}

\begin{table*}[!htbp]
    \centering
    \small
    \setlength{\tabcolsep}{3pt}
    \begin{tabular}{llccccccc}
    \toprule
    \textbf{Subset} & \textbf{Model} & \textbf{Pair} & \textbf{Cross NA} & \textbf{Cross Matched} & \textbf{Unm.\ (mirror)} & \textbf{Unm.\ (random)} & \textbf{Spread Avg} & \textbf{Spread Min} \\
    \midrule
    \multirow{15}{*}{Comp.}
        & \multirow{3}{*}{Llama-3.1-8B} & B+M & 28.47 & 37.62 & 33.11 & 26.51 & 29.70 & 24.83 \\
        &                               & B+T & 28.32 & 37.90 & 32.94 & 26.37 & 28.93 & 23.30 \\
        &                               & M+T & 24.12 & 36.54 & 31.16 & 23.45 & 24.07 & 23.30 \\
        & \multirow{3}{*}{Ministral-8B} & B+M & 31.63 & 34.77 & 33.38 & 29.83 & 34.27 & 28.89 \\
        &                               & B+T & 30.01 & 33.49 & 30.92 & 27.88 & 32.33 & 25.02 \\
        &                               & M+T & 25.67 & 31.21 & 28.20 & 22.88 & 26.95 & 25.02 \\
        & \multirow{3}{*}{Qwen2.5-7B} & B+M & 19.68 & 30.54 & 24.46 & 16.51 & 21.79 & 16.34 \\
        &                               & B+T & 21.30 & 30.66 & 26.28 & 19.73 & 22.40 & 17.55 \\
        &                               & M+T & 17.59 & 29.00 & 23.87 & 15.85 & 16.95 & 16.34 \\
        & \multirow{3}{*}{Qwen2.5-14B} & B+M & 36.95 & 42.87 & 38.38 & 25.41 & 38.23 & 34.89 \\
        &                               & B+T & 36.85 & 42.21 & 37.77 & 29.10 & 38.19 & 34.81 \\
        &                               & M+T & 34.28 & 41.83 & 37.21 & 29.75 & 34.85 & 34.81 \\
        & \multirow{3}{*}{Qwen3-8B} & B+M & 33.72 & 37.63 & 34.41 & 28.22 & 35.42 & 31.00 \\
        &                               & B+T & 32.58 & 36.82 & 32.94 & 27.88 & 34.62 & 29.40 \\
        &                               & M+T & 29.19 & 35.22 & 31.39 & 25.13 & 30.20 & 29.40 \\
    \midrule
    \multirow{15}{*}{Infer.}
        & \multirow{3}{*}{Llama-3.1-8B} & B+M & 17.26 & 29.12 & 22.49 & 14.45 & 20.70 & 12.91 \\
        &                               & B+T & 16.62 & 28.89 & 21.90 & 13.77 & 20.31 & 12.14 \\
        &                               & M+T & 11.46 & 23.96 & 19.38 & 14.90 & 12.52 & 12.14 \\
        & \multirow{3}{*}{Ministral-8B} & B+M & 13.04 & 17.98 & 15.49 & 10.91 & 16.94 &  8.83 \\
        &                               & B+T & 11.59 & 16.35 & 13.81 &  9.69 & 16.08 &  7.11 \\
        &                               & M+T &  7.65 & 12.59 & 10.33 &  6.30 &  7.97 &  7.11 \\
        & \multirow{3}{*}{Qwen2.5-7B} & B+M &  5.66 & 14.54 &  8.99 &  5.07 &  6.30 &  5.03 \\
        &                               & B+T &  5.75 & 13.63 &  9.19 &  5.30 &  6.57 &  5.57 \\
        &                               & M+T &  5.16 & 12.64 &  9.90 &  5.84 &  5.30 &  5.03 \\
        & \multirow{3}{*}{Qwen2.5-14B} & B+M & 16.80 & 24.32 & 20.90 & 11.50 & 20.99 & 11.96 \\
        &                               & B+T & 15.22 & 22.96 & 18.77 & 10.87 & 20.15 & 10.28 \\
        &                               & M+T & 10.51 & 19.47 & 15.76 &  8.24 & 11.12 & 10.28 \\
        & \multirow{3}{*}{Qwen3-8B} & B+M & 11.32 & 19.70 & 16.39 &  7.56 & 14.06 &  8.11 \\
        &                               & B+T & 10.55 & 17.35 & 13.68 &  7.34 & 13.11 &  6.20 \\
        &                               & M+T &  6.93 & 15.90 & 12.59 &  5.12 &  7.16 &  6.20 \\
    \bottomrule
    \end{tabular}
    \caption{2WikiMultiHopQA Cross protocol with Weakest Link references. Cross NA / Matched are per-pair accuracies (\%); Unmatched is split into (mirror) = mean of the two partial-gold mirror variants, and (random) = the random-distractor variant. Spread Avg / Min are the mean and minimum of the two constituent single-bucket Spread NA values.}
    \label{tab:2wiki-cross-weakest-link}
\end{table*}

\begin{table*}[!htbp]
    \centering
    \small
    \setlength{\tabcolsep}{4pt}
    \begin{tabular}{lcccccc}
    \toprule
    \textbf{Pair} & \textbf{Cross NA} & \textbf{Cross Matched} & \textbf{Unm.\ (mirror)} & \textbf{Unm.\ (random)} & \textbf{Spread Avg} & \textbf{Spread Min} \\
    \midrule
    B+M & 30.50 & 46.67 & 35.83 & 14.00 & 35.11 & 32.00 \\
    B+T & 31.67 & 44.67 & 38.00 & 25.17 & 35.50 & 32.78 \\
    M+T & 31.83 & 43.50 & 39.00 & 27.50 & 32.39 & 32.00 \\
    \bottomrule
    \end{tabular}
    \caption{Qwen2.5-32B-Instruct-GPTQ-Int8 Cross protocol with Weakest Link references. Cross NA / Matched are per-pair accuracies (\%); Unmatched is split into (mirror) = mean of the two partial-gold mirror variants, and (random) = random-distractor variant. Spread Avg / Min are the mean and minimum of the two single-bucket Spread NA values.}
    \label{tab:32b-cross-weakest-link}
\end{table*}

\subsubsection{NeoQA with Random-Timeline Distractors}

As discussed in \Cref{sec:robustness-verification}, NeoQA's original same-timeline distractors create context homogeneity that masks position bias. \Cref{tab:neoqa-random-timeline} shows the effect of replacing these with random-timeline distractors on Qwen2.5-7B-Instruct. Random distractors make the task easier overall, but a clear primacy bias emerges. Matched MFAI closes this gap, confirming that the attention steering mechanism is effective even when the underlying task topology is horizontal.

\begin{table}[ht]
    \centering
    \small
    \resizebox{\columnwidth}{!}{
    \begin{tabular}{llcccc}
    \toprule
    \textbf{Setting} & \textbf{Condition} & \textbf{Beg} & \textbf{Mid} & \textbf{Tail} & \textbf{Pos Bias (B--T)} \\
    \midrule
    \multirow{3}{*}{Same TL}   & NA       & 51.39\% & 51.04\% & 53.18\% & -1.79\% \\
                                & Matched  & 55.37\% & 51.49\% & 54.98\% & +0.40\% \\
                                & Unmatched& 51.77\% & 52.79\% & 55.77\% & -4.00\% \\
    \midrule
    \multirow{3}{*}{Random TL}   & NA       & 62.29\% & 58.76\% & 58.46\% & +3.83\% \\
                                & Matched  & 62.84\% & 62.59\% & 62.19\% & +0.65\% \\
                                & Unmatched& 62.24\% & 60.47\% & 61.84\% & +0.40\% \\
    \bottomrule
    \end{tabular}
    }
    \caption{NeoQA Spread Test: Same-Timeline vs.\ Random-Timeline distractors. Model: Qwen2.5-7B-Instruct. Averages across distances 1--5. TL = Timeline.}
    \label{tab:neoqa-random-timeline}
\end{table}

\subsubsection{Test-Time Compute Cost: Thinking vs.\ Non-Thinking}

The robustness gains of Qwen3-8B-Think come at a non-trivial inference cost. \Cref{tab:think-token-cost} reports the mean number of output tokens per example on NeoQA (same-timeline distractors) for Qwen3-8B in its non-thinking and thinking modes, pooled across all MFAI conditions and bucket positions. The thinking mode emits roughly $6\times$ more output tokens than its non-thinking counterpart on both Spread and Cross protocols, quantifying the budget required for the verification behavior discussed in the main text.

\begin{table}[ht]
    \centering
    \small
    \resizebox{\columnwidth}{!}{
    \begin{tabular}{lrrr}
    \toprule
    \textbf{Model} & \textbf{Spread} & \textbf{Cross} & \textbf{Ratio} \\
    \midrule
    Qwen3-8B (non-thinking) &  243.1 &  248.8 & 1.00$\times$ \\
    Qwen3-8B-Think          & 1485.3 & 1512.7 & 6.09$\times$ \\
    \bottomrule
    \end{tabular}
    }
    \caption{Mean output tokens per example on NeoQA (same-timeline distractors) for Qwen3-8B in non-thinking mode vs.\ Qwen3-8B-Think. Tokens counted over \texttt{reasoning\_content $+$ model\_response} using the Qwen3-8B tokenizer, pooled across all MFAI conditions and bucket positions (Spread: $n{=}24,120$; Cross: $n{=}36,180$).}
    \label{tab:think-token-cost}
\end{table}

\subsection{Attention Instruction Order Sensitivity}
\label{appx:order-sensitivity}

Our MFAI injects the target document indices as a natural-language list (e.g., ``Document $i$, Document $j$''). A potential concern is that the model could latch onto the specific ordering of this list rather than treating it as a set of cited positions, in which case reversing the index order inside the instruction would materially change the answer. We re-run every Cross Test cell with the indices inside the attention instruction reversed and compare against the original ordering. The test covers 96 cells in total (2 datasets $\times$ 2 models $\times$ 3 bucket-pairs $\times$ 2 local indices $\times$ 4 MFAI conditions). \Cref{tab:order-sensitivity} reports both the absolute percentage-point change $|\Delta|$ and the relative change $|\Delta|/\text{baseline}$, since a $4$\,pp shift means something very different on a $55\%$ baseline than on a $10\%$ baseline.

Across all 96 cells the mean absolute change is $1.15$\,pp with a median of $0.80$\,pp, and the median relative change is $2.52\%$. MuSiQue-Llama is essentially invariant (mean $0.69$\,pp, max relative $6.32\%$) and NeoQA-Llama has a median relative change of only $0.89\%$. The largest relative shifts are concentrated in MuSiQue-Qwen, where the baseline accuracy is close to random ($9.71$--$23.52\%$) so small absolute wiggles inflate in relative terms; all eight cells exceeding $10\%$ relative change fall within \texttt{unmatched\_*} control conditions rather than the \texttt{matched} cells that drive the main findings. We therefore treat the MFAI index list as order-agnostic throughout the paper.

\begin{table}[ht]
    \centering
    \small
    \setlength{\tabcolsep}{3pt}
    \resizebox{\columnwidth}{!}{
    \begin{tabular}{llcrrrrr}
    \toprule
    \textbf{Dataset} & \textbf{Model} & \textbf{Baseline} & \textbf{Mean $|\Delta|$} & \textbf{Med.\ $|\Delta|$} & \textbf{Max $|\Delta|$} & \textbf{Med.\ rel.} & \textbf{Max rel.} \\
    \midrule
    MuSiQue & Llama-3.1-8B & 18.22--33.71 & 0.69 & 0.60 & 1.77 &  2.59\% &  6.32\% \\
    MuSiQue & Qwen2.5-7B   &  9.71--23.52 & 1.25 & 1.24 & 3.13 &  7.20\% & 17.18\% \\
    NeoQA   & Llama-3.1-8B & 54.73--60.45 & 1.11 & 0.50 & 4.48 &  0.89\% &  8.19\% \\
    NeoQA   & Qwen2.5-7B   & 48.26--54.23 & 1.53 & 1.24 & 5.22 &  2.43\% & 10.82\% \\
    \midrule
    \multicolumn{2}{l}{\textit{Pooled (96 cells)}} & 9.71--60.45 & 1.15 & 0.80 & 5.22 & 2.52\% & 17.18\% \\
    \bottomrule
    \end{tabular}
    }
    \caption{Effect of reversing the MFAI index order on Cross Test accuracy. Each per-model row aggregates $n{=}24$ cells (3 bucket-pairs $\times$ 2 local indices $\times$ 4 MFAI conditions). \textbf{Baseline} is the range of original accuracies (percent) over the 24 cells. \textbf{$|\Delta|$} is the absolute percentage-point change in accuracy when the instruction indices are reversed, reported as the mean, median, and max over the 24 cells. \textbf{rel.} is the per-cell relative change $|\Delta_c|/\text{baseline}_c$ (as a percent); \textbf{Med.\ rel.} and \textbf{Max rel.} are the median and max of this quantity over the 24 cells.}
    \label{tab:order-sensitivity}
\end{table}

\subsection{Unmatched Instruction Variants and Detailed Analysis}
\label{appx:unmatched_variants}
\label{appx:detailed-unmatched}

To ensure that our \textbf{Unmatched MFAI} condition tests robustness against misleading signals rather than random noise, we generate adversarial indices by mirroring the local structure of the gold evidence. The specific variants for each protocol are:

\paragraph{Spread Test Variants:} When the gold set $\mathcal{G}$ resides entirely within one bucket (e.g., Beginning), we generate two unmatched variants:

\begin{enumerate}
    \item \textbf{Middle Mirror:} The instruction points to documents in the Middle bucket that share the same local indices as the gold documents.
    \item \textbf{Tail Mirror:} The instruction points to documents in the Tail bucket that share the same local indices.
\end{enumerate}

\paragraph{Cross Test Variants:} When $\mathcal{G}$ is split across two buckets (e.g., Beginning and Middle), we use three unmatched variants to average out the effects of partial correctness:

\begin{enumerate}
    \item \textbf{Partial Erroneous Mirror (Gold-1 correct):} The instruction correctly points to the gold document in the first bucket (Beginning) but points to a mirrored distractor in the non-gold bucket (Tail).
    \item \textbf{Partial Erroneous Mirror (Gold-2 correct):} The instruction correctly points to the gold document in the second bucket (Middle) but points to a mirrored distractor in the non-gold bucket (Tail).
    \item \textbf{Random Distractor:} The instruction points to two randomly selected documents within the non-gold bucket (Tail), ensuring no overlap with the gold indices.
\end{enumerate}

\paragraph{Per-Variant Results.}

\begin{figure*}[ht!]
    \centering
    \begin{subfigure}[b]{\textwidth}
        \centering
        \includegraphics[width=\textwidth]{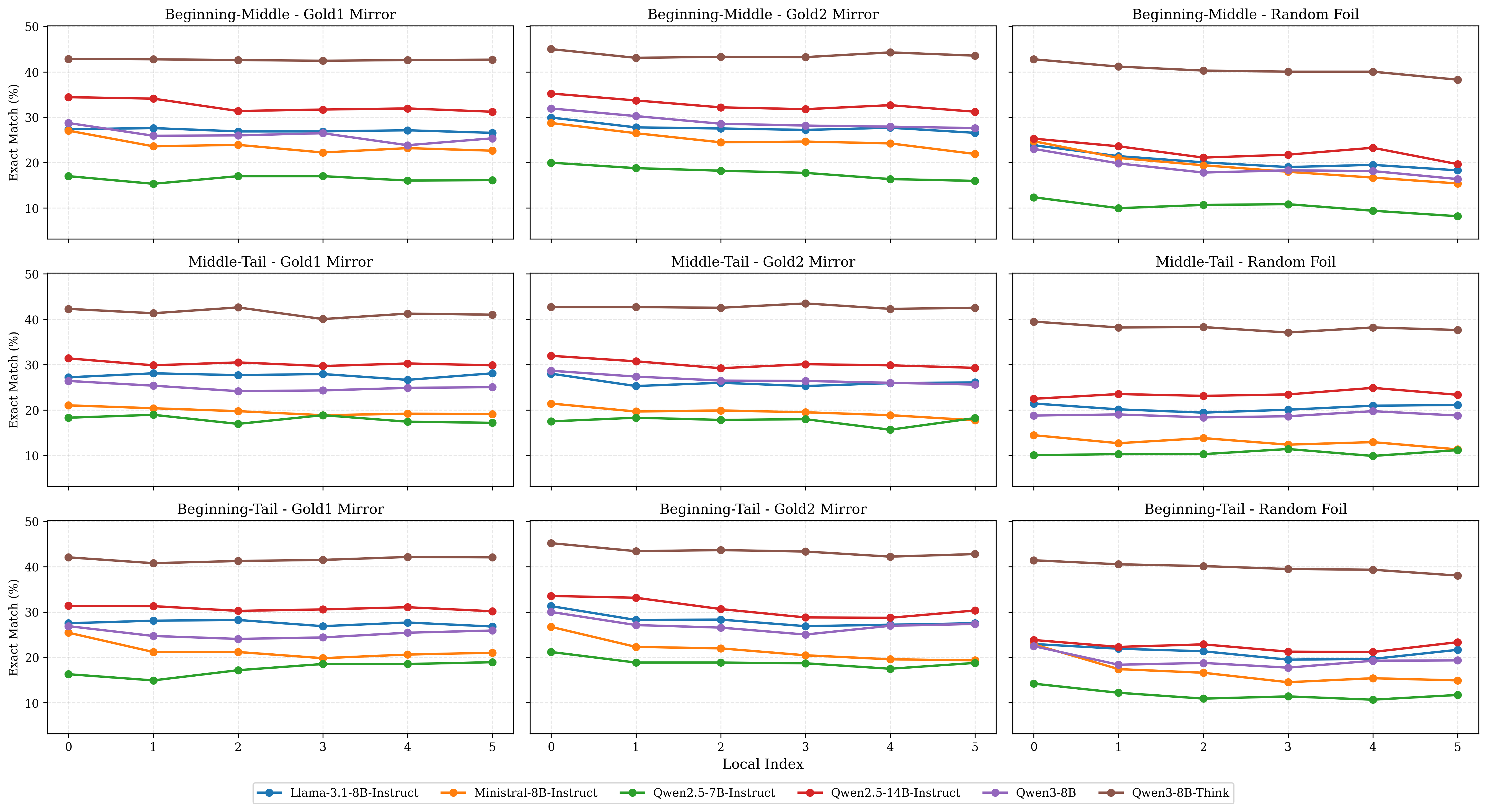}
        \caption{MuSiQue}
        \label{fig:cross_unmatched_variants_musique}
    \end{subfigure}

    \vspace{0.5em}

    \begin{subfigure}[b]{\textwidth}
        \centering
        \includegraphics[width=\textwidth]{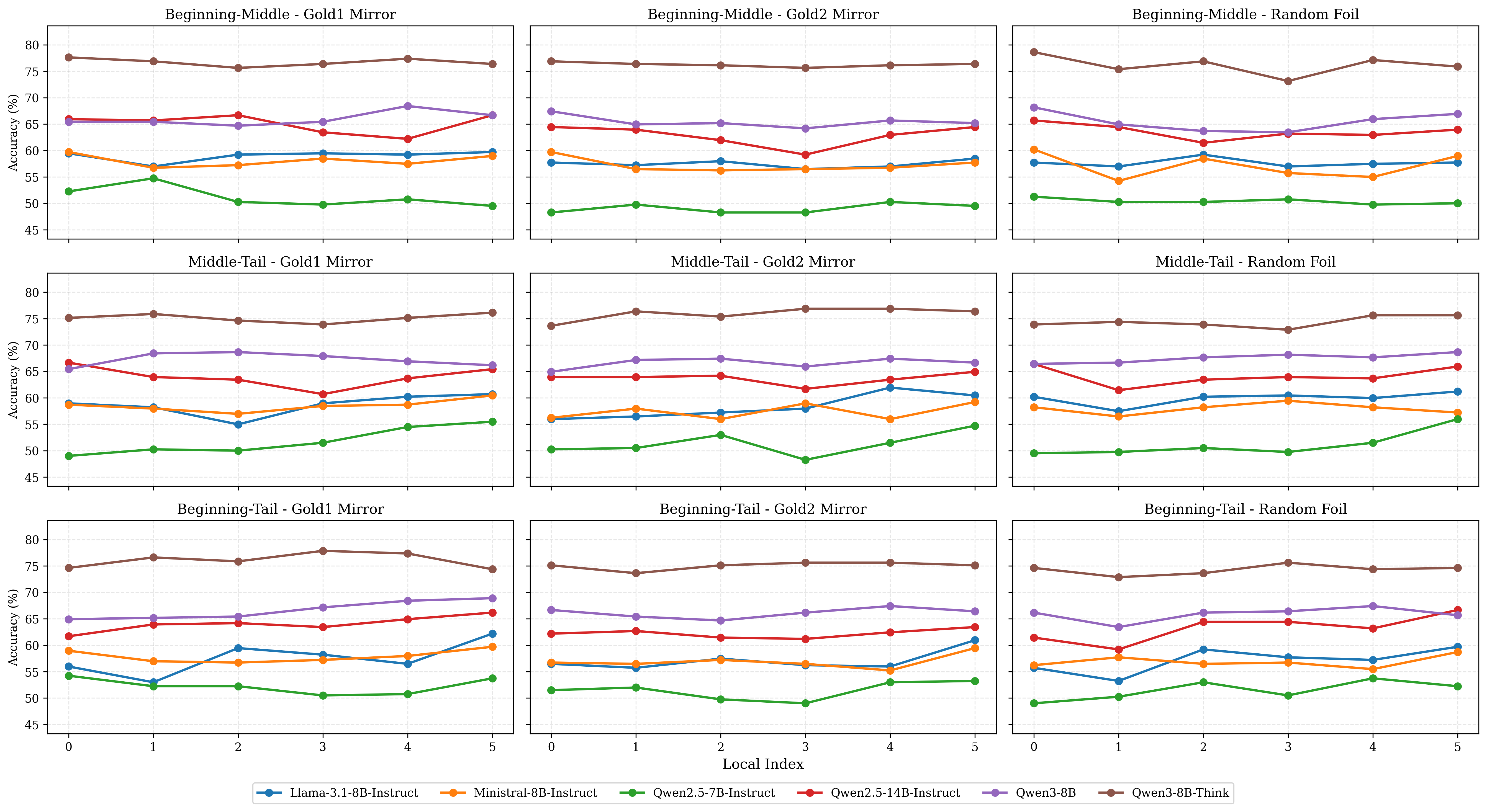}
        \caption{NeoQA}
        \label{fig:cross_unmatched_variants_neoqa}
    \end{subfigure}
    
    \caption{Plot of the unmatched variants of Cross Test for MuSiQue and NeoQA datasets. Each subplot shows the performance of models in one of the three unmatched variants at the selected bucket pair. The x-axis represents the local index within the selected bucket pair, and the y-axis represents the accuracy. Each row corresponds to a different bucket pair, with the first row showing the Beginning--Middle bucket pair, etc. Each column corresponds to a different unmatched variant.}
    \label{fig:cross_unmatched_variants}
\end{figure*}

\Cref{fig:cross_unmatched_variants} illustrates model performance on the unmatched variants of the Cross Test for MuSiQue and NeoQA. Comparing the variants in each bucket pair (rows), we observe that partially correct MFAI (gold1-mirror and gold2-mirror) remain helpful. The performance is better than with random misleading instructions and even exceeds the baseline (No MFAI), compared to solid lines in \Cref{fig:cross_test_local_idx}. This reinforces the recognition bottleneck hypothesis that model failures stem primarily from positional neglect due to attention deficiency, and even partially correct cues can restore the focus.

\subsection{Prompts}

We used a uniform prompt for MuSiQue across all models. For NeoQA, however, we selected the optimal instruction for each model using gold-only settings, following the methodology established by prior work~\citep{glockner2025neoqa}.

\subsubsection{The Prompt For MuSiQue}

The standard MuSiQue prompt:

\begin{tcolorbox}[
    colback=gray!3,
    colframe=gray!60,
    title=\textbf{MuSiQue Standard Prompt}
]
\begin{lstlisting}[language={}, basicstyle=\ttfamily\small, breaklines=true, columns=fullflexible]
In this task, you are presented with a question, and 18 documents that covers the answer to that question. Deduce your answer solely from the provided documents, avoiding any external data sources. Keep the answer short and concise, leave behind any irrelevant details.
{{ATTENTION_INSTRUCTION}}

Question: {{QUESTION}}

Documents:
{{DOCUMENTS_BLOCK}}

If the documents don't have the answer, set "is_answerable" to false in the output dictionary. If they do, set "is_answerable" to true and put the answer in "answer_content".

Please provide your answer in the following format:
{"is_answerable": true/false, "answer_content": "your answer here"}
\end{lstlisting}
\end{tcolorbox}

\paragraph{The Formatting of Document}
Each document is separated by line breaks.

\begin{tcolorbox}[
    colback=gray!3,
    colframe=gray!60,
    title=\textbf{MuSiQue Document Formatting}
]
\begin{lstlisting}[language={}, basicstyle=\ttfamily\small, columns=fullflexible]
Document 1: Title of Article
Text of the article content...

Document 2: Title of Article
Text of the article content...
...
\end{lstlisting}
\end{tcolorbox}

\subsubsection{The Prompt For NeoQA}

\begin{table}[h!]
\centering
\small
\begin{tabular}{ll}
\hline
\textbf{Model} & \textbf{Instruction Template} \\ \hline
Qwen3-8B & last-line-instructions-1 \\
Qwen2.5-7B-Instruct & last-line-instructions-2 \\
Qwen2.5-14B-Instruct & last-line-instructions-1 \\
Llama-3.1-8B-Instruct & last-line-instructions-2 \\
Ministral-8B-Instruct & last-line-instructions-1 \\
\end{tabular}
\caption{Mapping of models to NeoQA prompt instruction files.}
\end{table}

\begin{tcolorbox}[
    colback=gray!3,
    colframe=gray!60,
    title={Prompt: last-line-instructions-1},
    breakable, enhanced jigsaw
]
\begin{lstlisting}[
    language={},
    basicstyle=\ttfamily\scriptsize,
    breaklines=true,
    columns=fullflexible
]
Given the following news articles, the question, and the answer options, answer the question.
If the question cannot be answered with certainty based on the news articles, select the answer option "Unanswerable".

News Articles:
{{NEWS_ARTICLES}}

Question: {{QUESTION}}

Date of Question: {{DATE}}

Answer options:
{{ANSWERS}}

Select the answer option that correctly answers the question. If the question cannot be answered with certainty based on the news articles, choose "Unanswerable" (if it is one of the options). In the final line of your response, provide the number of the correct answer option using the format: "Answer: [answer number]" (for example, "Answer: X").
\end{lstlisting}
\end{tcolorbox}

\begin{tcolorbox}[
    colback=gray!3,
    colframe=gray!60,
    title={Prompt: last-line-instructions-2},
    breakable, enhanced jigsaw
]
\begin{lstlisting}[
    language={},
    basicstyle=\ttfamily\scriptsize,
    breaklines=true,
    columns=fullflexible
]
You will receive news articles, a question, a date on which the question is asked, and answer options.
Your task is to evaluate the articles, determine if they provide enough information to answer the question based on the date, and choose the correct answer.

News Articles:
{{NEWS_ARTICLES}}

Question: {{QUESTION}}

Date of Question: {{DATE}}

Answer options:
{{ANSWERS}}

**Instructions:**
1. **Analyze the news articles:**
- Carefully read all the news articles.
- Compare the information in the articles with the question.
- Check if the combined information from the articles confirms all the details required to answer the question.

2. **Select an Answer:**
- Choose the correct answer if all necessary details are provided.
- If the articles lack information or any important detail is missing, select the option for "Unanswerable".

3. **Submit your Answer**
- Select the answer option that correctly answers the question. If the question cannot be answered with certainty based on the news articles, choose "Unanswerable" (if it is one of the options). In the final line of your response, provide the number of the correct answer option using the format: "Answer: [answer number]" (for example, "Answer: X").
\end{lstlisting}
\end{tcolorbox}

\paragraph{News Articles Formatting ({{NEWS\_ARTICLES}})}
The documents are formatted using an XML-like structure that includes the title, date, and text content for each article. For each document in the context, the following structure is repeated, separated by two newlines:

\begin{tcolorbox}[
    colback=gray!3,
    colframe=gray!60,
    title=\textbf{Formatting of \texttt{\{\{NEWS\_ARTICLES\}\}}}
]
\begin{lstlisting}[language={}, basicstyle=\ttfamily\small, columns=fullflexible]
<article>
<title>Title of the Article</title>
<date>YYYY-MM-DD</date>
<text>Full text of the news article...</text>
</article>
\end{lstlisting}
\end{tcolorbox}

\paragraph{Answer Options Formatting ({{ANSWERS}})}
The multiple-choice options are formatted as a numbered list where each index is enclosed in square brackets. The answer options are provided as a list starting from index 1:

\begin{tcolorbox}[
    colback=gray!3,
    colframe=gray!60,
    title=\textbf{Formatting of \texttt{\{\{ANSWERS\}\}}}
]
\begin{lstlisting}[language={}, basicstyle=\ttfamily\small, columns=fullflexible]
[1] Text for option 1
[2] Text for option 2
[3] Text for option 3
...
[4] Unanswerable
\end{lstlisting}
\end{tcolorbox}

\end{document}